\documentclass{article}
\usepackage{caption}

\usepackage{microtype}
\usepackage{graphicx}
\usepackage{subfigure}
\usepackage{booktabs} 

\usepackage{hyperref}

\usepackage[accepted]{icml2025_arxiv}

\usepackage{amsmath}
\usepackage{amssymb}
\usepackage{mathtools}
\usepackage{amsthm}

\usepackage[capitalize,noabbrev]{cleveref}

\NewDocumentCommand{\heng}
{ mO{} }{\textcolor{red}{\textsuperscript{\textit{Heng}}\textsf{\textbf{\small[#1]}}}}

\theoremstyle{plain}

\theoremstyle{definition}

\theoremstyle{remark}

\usepackage[textsize=tiny]{todonotes}

\usepackage{color}
\usepackage{epsfig}
\usepackage{graphicx}

\usepackage{adjustbox}
\usepackage{array}
\usepackage{booktabs}
\usepackage{colortbl}
\usepackage{wrapfig}
\usepackage{hhline}
\usepackage{multirow}
\usepackage{wrapfig}
\usepackage{floatflt}
\usepackage{subfigure}

\usepackage{bm}
\usepackage{nicefrac}

\usepackage{changepage}
\usepackage{extramarks}
\usepackage{fancyhdr}
\usepackage{lastpage}
\usepackage{setspace}
\usepackage{soul}
\usepackage{xspace}

\usepackage{enumerate}
\usepackage{enumitem}  

\usepackage{makecell}

\usepackage{pifont} 

\usepackage[symbol]{footmisc}

\usepackage[most]{tcolorbox}

\newcommand{\ours}{Mobile-Agent-E\xspace}
\newcommand{\oursfull}{Mobile-Agent-E\xspace}
\newcommand{\dataset}{Mobile-Eval-E\xspace}
\newcommand{\datasetfull}{Mobile-Eval-E\xspace}

\definecolor{MyDarkBlue}{rgb}{0,0.08,0.8}
\definecolor{MyDarkGreen}{RGB}{45,155,45}
\definecolor{MyDarkRed}{rgb}{0.8,0.02,0.02}
\definecolor{MyOrange}{rgb}{1.0, 0.4, 0.2}
\definecolor{MyPurple}{RGB}{111,0,255}
\definecolor{MyRed}{rgb}{0.8,0.0,0.0}
\definecolor{MyGold}{rgb}{0.75,0.6,0.12}
\definecolor{MyDarkgray}{rgb}{0.66, 0.66, 0.66}

\newcommand{\ourtitle}{\ours: Self-Evolving Mobile Assistant for Complex Tasks}

\icmltitlerunning{\ourtitle}

\begin{document}

\twocolumn[
\icmltitle{\ourtitle}







\begin{icmlauthorlist}
\icmlauthor{Zhenhailong Wang \textsuperscript{*}}{uiuc}
\hspace{5pt}
\icmlauthor{Haiyang Xu \textsuperscript{*}}{alibaba}
\hspace{5pt}
\icmlauthor{Junyang Wang}{alibaba}
\hspace{5pt}
\icmlauthor{Xi Zhang}{alibaba}
\\
\vspace{3pt}
\icmlauthor{Ming Yan}{alibaba}
\hspace{5pt}
\icmlauthor{Ji Zhang}{alibaba}
\hspace{5pt}
\icmlauthor{Fei Huang}{alibaba}
\hspace{5pt}
\icmlauthor{Heng Ji \textsuperscript{*}}{uiuc}
\end{icmlauthorlist}

\icmlaffiliation{uiuc}{University of Illinois Urbana-Champaign}
\icmlaffiliation{alibaba}{Alibaba Group}

\icmlcorrespondingauthor{Zhenhailong Wang}{wangz3@illinois.edu}
\icmlcorrespondingauthor{Haiyang Xu}{shuofeng.xhy@alibaba-inc.com}
\icmlcorrespondingauthor{Heng Ji}{hengji@illinois.edu}

\icmlkeywords{Machine Learning, ICML}

\vskip 0.4in




{
\begin{center}
    \centering
    \captionsetup{type=figure}
    \vspace{-0.5cm}
    \includegraphics[width=0.99\textwidth]{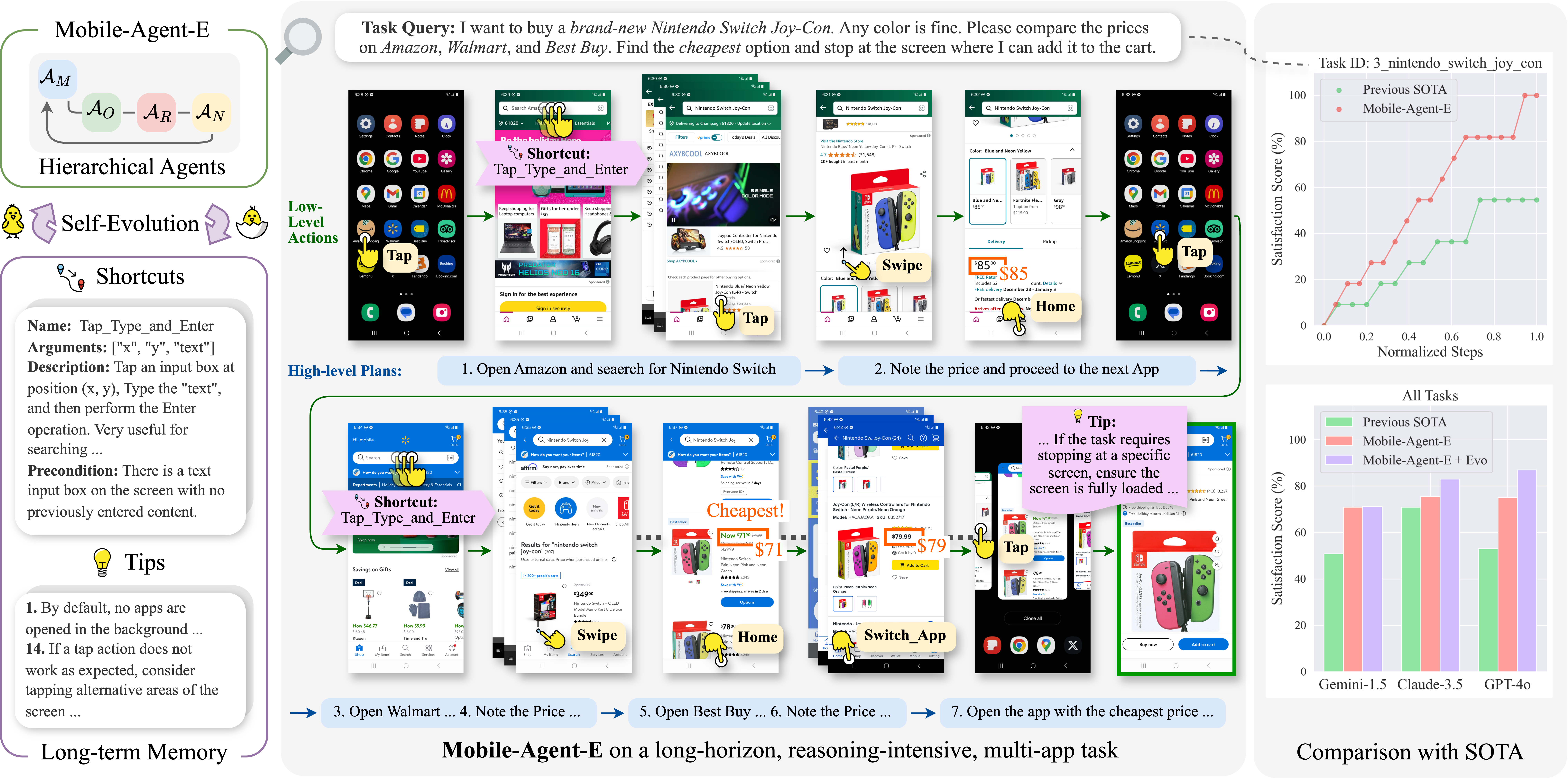}
    \vspace{-5pt}
    \caption{We propose \ours, a novel hierarchical multi-agent mobile assistant that outperforms previous state-of-the-art approaches~\cite{yang2023appagent, wang2024mobile, wang2024mobile2} on complex real-world tasks. \ours disentangles high-level planning and low-level action decision with dedicated agents. Equipped with a newly introduced self-evolution module that learns general \textit{Tips} and reusable \textit{Shortcuts} from past experiences, \ours demonstrates further improvements in both performance and efficiency.
    }
    \label{fig:teaser}
    \vspace{0.1cm}
\end{center}%
}


]



\printAffiliationsAndNotice{}  


\begin{abstract}

Smartphones have become indispensable in modern life, yet navigating complex, multi-step tasks on mobile devices often remains frustrating and time-consuming. Recent advancements in large multimodal model (LMM)-based mobile agents have demonstrated the ability to perceive and act in mobile environments on behalf of users. However, current approaches face significant limitations: they fall short in addressing real-world human needs, struggle with reasoning-intensive and long-horizon tasks, and lack mechanisms to learn and improve from prior experiences. 
To overcome these challenges, we introduce \textbf{\ours}, a hierarchical multi-agent framework capable of self-evolution through past experience.
By ``hierarchical,'' we mean an explicit separation of high-level planning and low-level action execution. 
The framework comprises a Manager, responsible for devising overall plans by breaking down complex tasks into subgoals, and four subordinate agents---Perceptor, Operator, Action Reflector, and Notetaker---which handle fine-grained visual perception, immediate action execution, error verification, and information aggregation, respectively.
\ours also features a novel self-evolution module which maintains a persistent long-term memory comprising \textit{Tips} and \textit{Shortcuts}.
Tips are general guidance and lessons learned from prior tasks on how to effectively interact with the environment.
Shortcuts are reusable, executable sequences of atomic operations tailored for specific subroutines.
The inclusion of Tips and Shortcuts facilitates continuous refinement of task performance and efficiency.
Alongside this framework, we introduce \textbf{\dataset}, a new benchmark featuring complex real-world mobile tasks requiring long-horizon, multi-app interactions.
Empirical results show that \ours achieves a 22\% absolute improvement over previous state-of-the-art approaches across three foundation model backbones.
Additionally, we provide a comprehensive analysis of the impact of our self-evolution mechanism and suggest directions for future work.
Code and data are publicly available for research purposes at \url{https://x-plug.github.io/MobileAgent}.

\end{abstract}

\section{Introduction}
\label{sec:intro}

Smartphones have become integral to our daily lives, transforming the way we connect, work, and find entertainment. Yet, the average 4.5 hours people spend on their phones daily\footnote{\url{https://explodingtopics.com/blog/smartphone-usage-stats}} often includes moments of frustration. Tedious tasks, such as deal hunting across multiple apps or gathering scattered information from various websites, often make us wish for a smart mobile assistant to ease these burdens.
Recent advancements in large multimodal models (LMMs)~\cite{openai2024gpt4o, claude35, team2024gemini15} have led to the emergence of LMM-based GUI agents~\cite{wang2024guisurvey, nguyen2024gui} capable of perceiving and acting in the Web, PC, and mobile environments on behalf of human users. 
Despite these initial successes, current research on mobile agents~\cite{wang2024mobile, yang2023appagent, wang2024mobile2, li2024appagentv2} has yet to fully address the challenges of real-world mobile tasks. We identify two key limitations below.

First, we observe a significant gap between the capabilities of current mobile agents and the demands of real-world scenarios. While existing mobile agent tasks are typically short, straightforward, and goal-oriented, such as ``Navigate to a nearby gas station''~\cite{wang2024mobile2}, tasks that better reflect actual human needs are far more complex.
These tasks often require a combination of (1) intensive reasoning to address multiple constraints, such as balancing various factors or criteria; (2) long-horizon planning, which may involve a lengthy sequence of steps across multiple apps; and (3) exploration, where the instructions can be vague and require active information gathering rather than following a fixed trajectory.
For instance, as shown in Figure~\ref{fig:teaser}, online shopping often involves navigating across different apps to compare prices and find the best deal.
Furthermore, the highly dynamic nature of mobile environments, characterized by pop-up advertisements and frequently changing app layouts, poses additional challenges in tackling these complex real-world tasks.

Second, unlike humans, who quickly adapt and become proficient with new devices or apps, current mobile agents lack the ability to learn from prior experiences. For example, when a human user first opens an app like Maps, it may take some trial and error to understand the layout and successfully perform a search. However, with each interaction, the user learns, becoming faster and more accurate the next time. 
In contrast, existing mobile agents treat every task as if it were their first attempt, allocating the same computational resources at each step and repeating the same mistakes, regardless of how many times they perform the same task. This inability to accumulate knowledge and refine actions from past experiences severely limits their ability to handle the aforementioned complex, long-horizon tasks, where subroutines such as searching and creating notes are often shared across different objectives.

To address these limitations, we propose \textbf{\oursfull}, a \textbf{hierarchical multi-agent framework} capable of \textbf{self-evolution} through past experiences.
\oursfull explicitly disentangles high-level planning---such as decomposing a task into smaller subgoals---from low-level actions, which involves determining specific actions and their parameters (e.g., \texttt{tap(x,y)}).
The framework is structured with a Manager, responsible for creating overall plans, and four subordinate agents---Perceptor, Operator, Action Reflector, and Notetaker---that handle fine-grained visual perception, action decision, outcome verification, and information aggregation, respectively.
This hierarchical design significantly enhances long-term planning and improves error recovery in complex tasks. 
Figure~\ref{fig:teaser} shows an overview of \ours on a challenging online shopping task requiring multi-step reasoning and interaction across three different apps.

\oursfull also features a self-evolution module, which includes a persistent long-term memory and two Experience Reflectors.
We define two types of critical knowledge that are continuously updated in the long-term memory across tasks: \textbf{Tips}---general guidance on effective interactions and lessons learned from previous trail-and-error experiences---and \textbf{Shortcuts}---reusable, executable functions that contains sequences of atomic operations tailored to efficiently complete recurring subroutines under specific preconditions.
After completing each task, the Experience Reflectors are triggered to update the Tips and propose new Shortcuts based on the interaction history. 
These are then fed to the Manager and Operator, enabling improved planning and action decision-making in future tasks.
This design draws inspiration from human cognitive science, where Tips are akin to the lessons encoded in episodic memory~\cite{tulving2002episodic}, which involves recalling specific past experiences and using them to inform future decisions, while Shortcuts resemble procedural knowledge that facilitates the efficient and often subconscious execution of well-practiced tasks~\cite{squire1996structure, anderson1982acquisition}.
An example of Shortcuts and Tips is provided in Figure~\ref{fig:teaser}.

To address the limitation of existing mobile benchmarks, which mainly include short-horizon and straightforward tasks with already saturated performance, we introduce a new benchmark, \textbf{\datasetfull}, designed to evaluate complex, real-world tasks.
\dataset features more than \textit{twice} the number of expected operations per task compared to previous benchmarks~\cite{wang2024mobile,yang2023appagent,wang2024mobile2} and incorporating a significantly higher proportion of tasks requiring \textit{multi-app} interactions. 
Accompanying the benchmark, we introduce a new evaluation metric called the Satisfaction Score to address the challenge posed by real-world tasks that often lack a binary success flag or a ground truth trajectory. 
This metric is computed based on human-written rubrics that account for both milestone completion, such as ``opened Maps,'' and exploratory behaviors, such as ``viewed more than one review.'' This approach offers a reliable measure of agent performance aligned with human preferences.
We further propose a Satisfaction Score vs Steps (SSS) curve to better evaluate and visualize the efficiency of mobile agents.
\dataset sets a high standard of difficulty, with prior state-of-the-art methods achieving only about 50–70\% of human satisfaction.

Empirical results show that \ours achieves an average absolute gain of 22.1\% over previous state-of-the-art approaches across three different foundation model backbones.
\ours also demonstrates promising self-evolution behavior in both performance and efficiency, resulting in a 6.5\% absolute improvement compared to no evolution.
The incorporation of Shortcuts further reduces the computational overhead, achieving speeds comparable to prior models while delivering significantly better performance. Additionally, we provide a comprehensive analysis of various aspects of self-evolution’s impact and outline directions for future work.
\begin{figure*}[ht]
\vskip 0.2in
\begin{center}
\centerline{\includegraphics[width=0.8\textwidth]{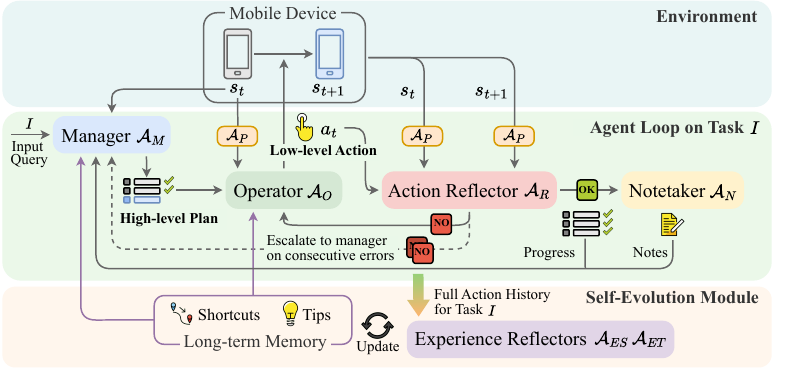}}
\vspace{-5pt}
\caption{An overview of the \ours framework, where the Manager, Perceptor ($\mathcal{A}_P$), Operator, Action Reflector, and Notetaker are involved in the main agent loop for each task, while two Experience Reflectors contribute to updating long-term memory across tasks. Decision-making at each step is disentangled into high-level planning by the Manager and low-level actions by the Operator. The Action Reflector verifies the outcome of each action, tracks progress, and provides error feedback. The Notetaker aggregates important information during navigation. A detailed example illustrating one step in the agent loop and the self-evolution process is presented in Figures~\ref{fig:example_breakdown} and \ref{fig:evolve_breakdown}.
}
\label{fig:agent_overview}
\end{center}
\vskip -0.2in
\end{figure*}

\begin{figure*}[ht]
\vskip 0.2in
\begin{center}
\centerline{\includegraphics[width=0.9\textwidth]{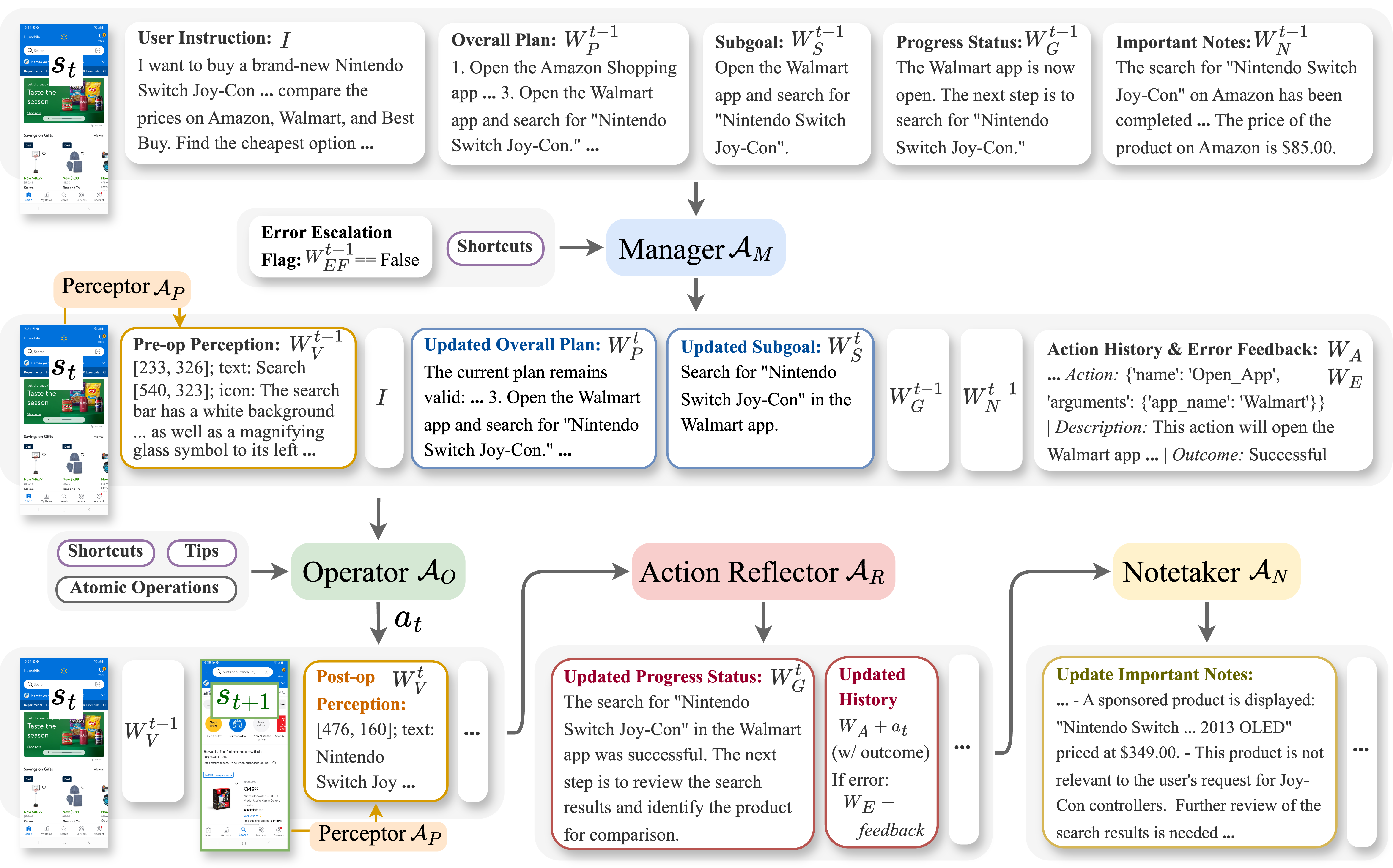}}
\vspace{-5pt}
\caption{A detailed breakdown of one inference step $t$ with \ours, showing the inputs and outputs of each agent. Omitted information indicates no change.
}
\label{fig:example_breakdown}
\end{center}
\vskip -0.2in
\end{figure*}

\section{\oursfull{}}
\label{sec:method}
\begin{table}[t]
\caption{Notation definitions.}
\label{tab:notation}
\vskip 0.1in
\begin{center}
\begin{small}
{
\centering
\setlength{\tabcolsep}{4pt} 
\begin{tabular}{ll}
\toprule
\textbf{Notation} & 
\textbf{Description} \\
\midrule
\multicolumn{2}{l}{\textit{Environment}} \\
\midrule
$I$ & Input task query \\
$a^t$ & Action\footnotemark{} at time $t$ \\
$s^t$ & Phone state (screenshot) at time $t$ \\
\midrule
\multicolumn{2}{l}{\textit{Agents}} \\
\midrule
$\mathcal{A}_P$ & Perceptor\\
$\mathcal{A}_M$ & Manager\\
$\mathcal{A}_O$ & Operator\\
$\mathcal{A}_R$ & Action Reflector\\
$\mathcal{A}_N$ & Notetaker\\
$\mathcal{A}_{ES}$ & Experience Reflector for Shortcuts\\
$\mathcal{A}_{ET}$ & Experience Reflector for Tips\\
\midrule
\multicolumn{2}{l}{\textit{Working Memory}} \\
\midrule
$W_{V}^t$ & Visual perception result at time $t$\\
$W_{P}^t$ & Overall plan (decomposed subgoals) at time $t$\\
$W_S^t$ & Current subgoal at time $t$\\
$W_G^t$ & Progress status at time $t$\\
$W_N^t$ & Important notes at time $t$\\
$W_{EF}^t$ & Error Escalation Flag at time $t$\\
$\mathbf{W_{A}}$ & Action history with outcome status\\
$\mathbf{W_{E}}$ & Error history with feedback\\
\midrule
\multicolumn{2}{l}{\textit{Long-term Memory}} \\
\midrule
$L_{S}$ & Shortcuts\\
$L_{T}$ & Tips\\
\bottomrule
\end{tabular}
}
\end{small}
\end{center}
\vspace{-15pt}
\end{table}

\footnotetext{$a_t$ can represent either a single atomic operation or a sequence of atomic operations if performing a Shortcut.} 

Figure~\ref{fig:agent_overview} provides an overview of \ours. A summary of the notation definitions is presented in Table~\ref{tab:notation}.
We detail the hierarchical multi-agent framework (\S\ref{subsec:agent_framework}) and the self-evolution module (\S\ref{subsec:self_evo_module}) in \ours below.

\subsection{Hierachical Multi-Agent Framework}
\label{subsec:agent_framework}

Figure~\ref{fig:example_breakdown} provides a detailed breakdown of the main agent loop with concrete examples.
Except for the Perceptor, all reasoning agents are instantiated from a frozen large multimodal model (LMM), such as GPT-4o~\cite{openai2024gpt4o}. The inputs and outputs of each agent are detailed as follows.

\paragraph{Manager ($\mathcal{A}_M$): High-level planning.} The Manager focuses on devising high-level plans to achieve the user’s requests. At each step, the Manager checks the input query $I$, the current screenshot $s_t$, the previous overall plan $W_P^{t-1}$, the previous subgoal $W_S^{t-1}$, the progress status $W_G^{t-1}$, available Shortcuts from long-term memory $L_S$, and any recorded important notes $W_N^{t-1}$ to provide an updated overall plan $W_P^t$ and identify the next immediate subgoal $W_S^t$ to achieve. Note that the Manager does not condition on the fine-grained perception results from the Perceptor, as it is not necessary and can add noise to high-level planning. 

\begin{small}
\begin{align}
    W_P^t, W_S^t = &\mathcal{A}_M(I, s_t, W_P^{t-1}, W_S^{t-1}, W_G^{t-1}, W_N^{t-1}, L_S) \notag \\
        &\text{\quad if $t \geq 0$ and $W_{EF}^{t-1}$ == False}\\
    W_P^t, W_S^t = &\mathcal{A}_M(I, s_t, W_P^{t-1}, W_S^{t-1}, W_G^{t-1}, W_N^{t-1}, L_S, \notag \\  
    &\quad \mathbf{W_{E}}[-k:]) \text{\quad if $t \geq k$ and $W_{EF}^{t-1}$ == True}
\end{align}
\end{small}

Additionally, when the model is potentially stuck in an error loop, that is, observing $k$ consecutive failed actions (e.g., $k=2$) reported by the Action Reflector, a special Error Escalation Flag $W_{EF}^{t-1}$ will be raised to the Manager. In such cases, the Manager will be prompted with additional information about the recent errors $\mathbf{W_{E}}[-k:]$ and asked to determine how to address the error from a higher-level perspective---such as refining the overall plan or adjusting the current subgoal to rectify the issue.
In other cases, when an error first occurs, the Operator will attempt to address it before escalating the issue to the Manager.
A concrete example of how the error escalation can help recovering from errors can be found in Figure~\ref{fig:error_escalation}.


\paragraph{Perceptor ($\mathcal{A}_P$): Fine-grained visual perception.}
The Perceptor aims to detect and recognize rich information about the current phone state, such as icons and text.
We use a purely vision-based perception module that does not rely on the underlying XML file, following \cite{wang2024mobile2}. The Perceptor consists of three main components: an OCR model, an icon grounding model, and an icon captioning model. Given a screenshot $s_t$ at time $t$, the Perceptor generates a fine-grained list of texts and icons, along with their corresponding coordinates $W_V^t$. Note that we still provide the original screenshot image to subsequent reasoning agents as a holistic visual context.

\vspace{-10pt}
\begin{small}
\begin{equation}
    W_V^t = \mathcal{A}_P (s_t)
\end{equation}
\end{small}
\vspace{-10pt}

\paragraph{Operator ($\mathcal{A}_O$): Low-level action decisions.} The Operator decides which concrete action to perform based on the input query $I$, the overall plan $W_P^t$ and current subgoal $W_S^{t}$ from the Manager, the previous progress status $W_G^{t-1}$, the important notes $W_N^{t-1}$, along with a history of the latest $m$ actions $\mathbf{W_A}[-m:]$ and errors $\mathbf{W_{E}}[-m:]$.\footnote{We empirically set $m=5$ in our experiments.} 
The action history includes both the action and its outcome (success or failure). The Operator is explicitly prompted to rectify errors if it observes unresolved failures in the history. The Operator also considers the \textit{Tips} as guidance from the long-term memory, which can be self-evolved from past experiences. 
To enable accurate generation of the action parameters, e.g., the (x,y) coordinates on the screen for tapping, we also provide the Operator with the fine-grained perception results $W_V^t$ from the Perceptor along with the screenshot $s_t$.

\vspace{-10pt}
\begin{small}
\begin{align}
    a_t = \mathcal{A}_O &(I,s_t,W_V^t,W_P^t,W_S^t, W_G^t,W_N^t,\notag \\&\quad \mathbf{W_A}[-m:],\mathbf{W_E}[-m:],L_S,L_T)
\end{align}
\end{small}
\vspace{-10pt}

The output of the Operator is the next action $a_t$ to perform. The action space is defined to contain not only \textit{Atomic Operations} but also \textit{Shortcuts}, which can evolve through tasks.
The atomic operations include
\texttt{Open\_App},
\texttt{Tap},
\texttt{Swipe},
\texttt{Type},
\texttt{Enter},
\texttt{Switch\_App},
\texttt{Back},
\texttt{Home},
and \texttt{Wait}.
The full descriptions of the atomic operations can be found in Table~\ref{tab:atomic_operation_space}.
We detail the definitions and examples of \textit{Shortcuts} and \textit{Tips} in \S\ref{subsec:self_evo_module}.
\paragraph{Action Reflector ($\mathcal{A}_R$): Reflection on the action outcome.} The Action Reflector checks the screenshots before ($s_t$) and after ($s_{t+1}$) of an action ($a_t$) to verify if the previous action achieves the expected outcome.
We define three types of outcomes for an action: \textbf{A.} Successful or partially successful: the result of the last action meets the expectation;\footnote{Some actions may need multiple repetitions to fulfill the expectation, for example, swipe up to find reviews. Thus, we include partially successful as meeting the expectation.} \textbf{B.} Failed: the last action results in a wrong page; and \textbf{C.} Failed: the last action produces no changes.
After identifying the outcome, if the outcome is A, the Action Reflector updates the action history $\mathbf{W_A}[t]$ as well as the progress status $W_G^t$. If the outcome is B or C, the Action Reflector additionally provides a description of the error and suggests potential reasons and solutions in $\mathbf{W_E}[t]$.

\begin{small}
\begin{align}
    W_V^{t+1} = \mathcal{A}_P (s_{t+1}) &\text{\;\; \# run Perceptor on $s_{t+1}$} \\
    \mathbf{W_A}[t], \mathbf{W_E}[t], W_G^t = \mathcal{A}_R &(I,s_t,W_V^t, s_{t+1},W_V^{t+1},\notag \\ 
    & a_t,W_S^{t},W_G^{t-1},) 
\end{align}
\end{small}

\paragraph{Notetaker ($\mathcal{A}_N$): Information aggregation.} In complex mobile tasks, we often need to keep track of important notes during exploration, such as the price of a product or the phone number of a restaurant. The Notetaker is dedicated to extracting and aggregating task-relevant information $W_N^{t}$ after each step, based on the input query $I$, overall plan $W_P^t$, current subgoal $W_S^t$, current progress $W_G^t$, fine-grained screen perception $W_V^{t+1}$ after executing the action, and existing notes $W_N^{t-1}$.

\vspace{-5pt}
\begin{small}
\begin{align}
    W_N^t = \mathcal{A}_N (I, s_{t+1}, W_V^{t+1}, W_P^t, W_S^t, W_G^t, W_N^{t-1})
\end{align}
\end{small}
\vspace{-5pt}

\begin{figure*}[ht]
\vskip 0.2in
\begin{center}
\centerline{\includegraphics[width=0.9\textwidth]{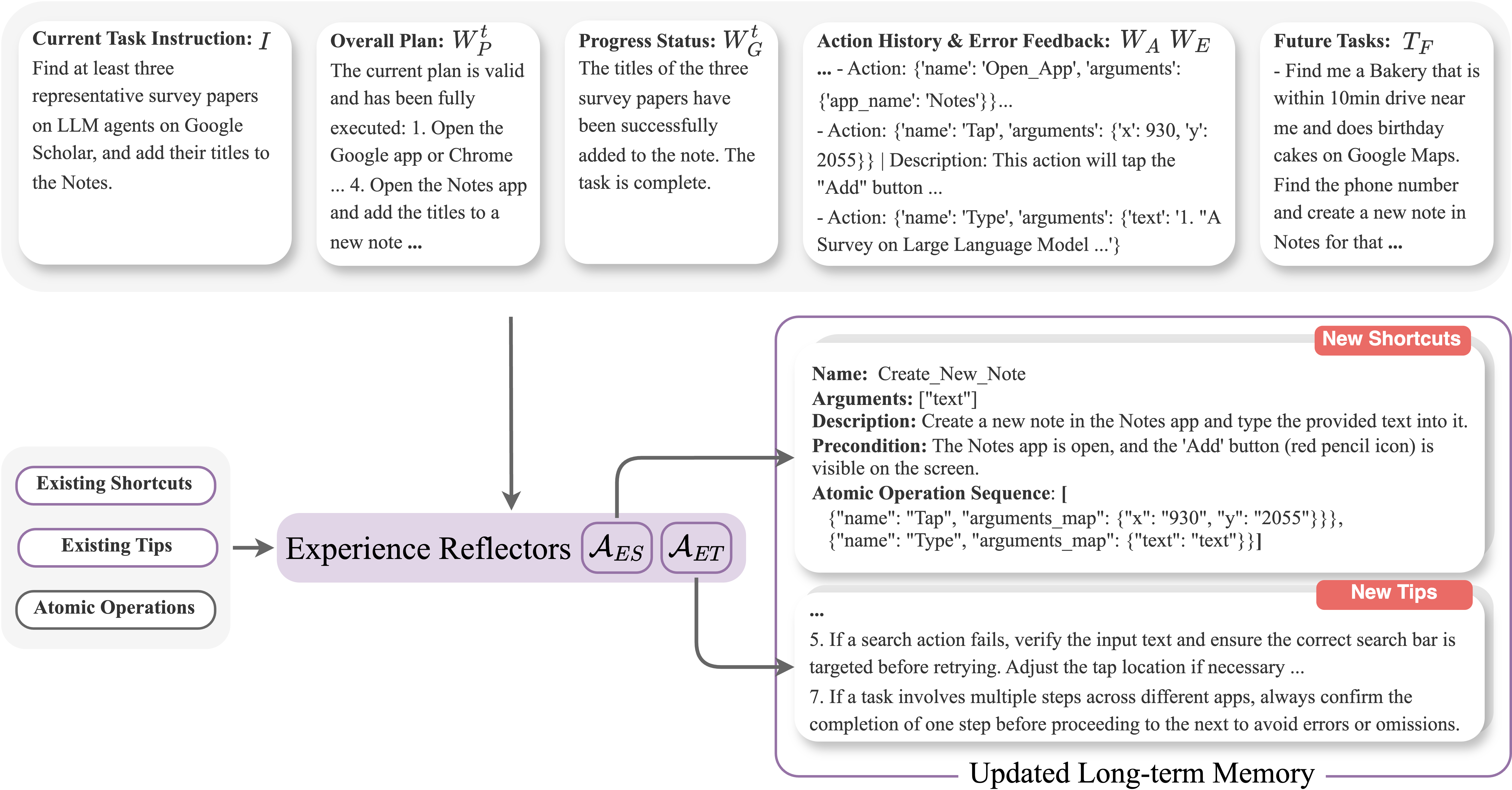}}
\vspace{-5pt}
\caption{Illustration of the inputs and outputs to the Experience Reflectors for a single self-evolution step, including a concrete example of the newly generated Shortcuts and Tips.
}
\label{fig:evolve_breakdown}
\end{center}
\vskip -0.2in
\end{figure*}
\subsection{Self-Evolution Module}
\label{subsec:self_evo_module}
Inspired by how humans become increasingly effective and efficient in operating smartphones, we maintain a \textbf{long-term memory} that \textbf{persists across tasks} and leverage two dedicated agents to reflect on past experiences. The long-term memory contains two important types of knowledge to evolve upon, \textbf{Tips} and \textbf{Shortcuts}, aiming to improve both the \textbf{performance} and \textbf{efficiency} of the agent.
Figure~\ref{fig:evolve_breakdown} provides a detailed breakdown of one self-evolution step.

\textbf{Tips} ($L_T$) are defined as general guidance on effective interactions and lessons learned from previous trial-and-error experiences. Tips resemble episodic memory~\cite{tulving2002episodic}, which enables humans to recall past experiences and apply insights to future decisions.

\textbf{Shortcuts} ($L_S$) are defined as reusable, executable functions composed of sequences of atomic operations tailored for recurring subroutines. Shortcuts are akin to procedural knowledge, which allows humans to perform well-practiced tasks efficiently and often subconsciously~\cite{squire1996structure, anderson1982acquisition}.
Due to the highly dynamic nature of the mobile environment, a Shortcut may only be applicable in certain states. For instance, the ``\texttt{Tap\_Type\_and\_Enter}'' Shortcut is usable only when the current screen has a text input box. To address this, we explicitly include a \textbf{precondition} in the definition of a Shortcut and require the Operator to verify that the current state satisfies the precondition before using the Shortcut. 
The arguments of a Shortcut have a unique one-to-one mapping to the arguments of its atomic operations.

When the self-evolution module is enabled, we leverage two Experience Reflectors, $\mathcal{A}_{ES}$ and $\mathcal{A}_{ET}$, to update the Tips and Shortcuts at the end of each task. The Experience Reflectors are also instantiated from frozen large multimodal model such as GPT-4o.
Let the final time step of a task be $t=\tau$. The input to the Experience Reflectors includes the input query $I$, the final overall plan $W_P^\tau$, the final progress status $W_G^\tau$, the entire action history $\mathbf{W_A}$ and error history $\mathbf{W_E}$, the existing Shortcuts $L_S$ and Tips $L_T$, and a list of future tasks $T_F$ (if provided). The outputs consist of newly generated Shortcuts in a predefined JSON format and updated Tips in natural language.
Figures~\ref{fig:full_shortcuts} and \ref{fig:full_tips} shows a full list of generated Shortcuts and Tips by \ours.

\begin{small}
\begin{align}
    L_T &= \mathcal{A}_{ET} (I, W_P^\tau, W_G^\tau, \mathbf{W_A}, \mathbf{W_E}, T_F, L_T)\\
    L_S &= \mathcal{A}_{ES} (I, W_P^\tau, W_G^\tau, \mathbf{W_A}, \mathbf{W_E}, T_F, L_S)
\end{align}
\end{small}

The updated Tips and Shortcuts are then utilized by the Manager and the Operator in the subsequent task, facilitating evolution in both high-level planning and low-level action decisions.
\section{Experiments}
\label{sec:experiments}
We perform a dynamic evaluation, evaluating the models in real-time and on actual devices---following previous work~\cite{wang2024mobile, yang2023appagent, wang2024mobile2}. Specifically, we use the Android Debug Bridge (ADB) to control an Android phone\footnote{A Samsung Galaxy A15 is used for all experiments.} and perform human evaluation on the recorded screenshots and action histories.

\begin{table}[t]
\caption{Comparison with existing dynamic evaluation benchmarks on real devices. \dataset emphasizes long-horizon, complex tasks that require significantly more operations and a wider variety of apps.
}
\label{tab:benchmark}
\vspace{-5pt}
\begin{center}
\begin{small}
\resizebox{0.49\textwidth}{!}
{
\centering
\setlength{\tabcolsep}{4pt} 
\begin{tabular}{lccccc}
\toprule
Benchmark & 
\#Tasks & 
\begin{tabular}[c]{@{}c@{}} \#Multi-App\\Tasks \end{tabular} & 
\begin{tabular}[c]{@{}c@{}} \#Apps \end{tabular} & 
\begin{tabular}[c]{@{}c@{}} Avg \#\\Ops \end{tabular} & 
\begin{tabular}[c]{@{}c@{}} Total \#\\Ops \end{tabular}\\
\midrule
Mobile-Eval        & 33 & 3  & 10 & 5.55 & 183 \\
Mobile-Eval-v2 & 44 & 4  & 10 & 5.57 & 245 \\
AppAgent             & \textbf{45} & 0  & 9  & 6.31 & 284 \\
\midrule
\dataset{}                  & 25 & \textbf{19} & \textbf{15} & \textbf{14.56} & \textbf{364} \\
\bottomrule
\end{tabular}
}
\end{small}
\end{center}
\vskip -0.15in
\end{table}

\subsection{A More Challenging Benchmark: \dataset}
Existing dynamic benchmarks~\cite{wang2024mobile, yang2023appagent, wang2024mobile2} primarily focus on short-horizon, straightforward tasks, where the performance has already saturated. To address this limitation, we propose a new and challenging benchmark, \textbf{\dataset}, which emphasizes reasoning-intensive, long-horizon, multi-app tasks. \dataset comprises 25 manually crafted tasks spanning 5 real-world scenarios: ``Restaurant Recommendation'', ``Information Searching'', ``Online Shopping'', ``What’s Trending'', and ``Travel Planning''. As shown in Table~\ref{tab:benchmark}, \dataset significantly surpasses previous benchmarks in complexity, featuring more than $\mathbf{2\times}$ the number of expected operations per task and a greater total number of operations. Most tasks in existing benchmarks can be viewed as specific subgoals in \dataset. Additionally, \dataset encompasses a broader range of Apps, with 76\% of the tasks requiring interactions with multiple Apps---compared to less than 10\% in previous benchmarks. In \S\ref{sec:experiments}, we demonstrate that this benchmark presents a substantial challenge for existing state-of-the-art models. The full set of task queries can be found in Appendix Table~\ref{tab:all_benchmark_tasks}. Due to the long-horizon nature of the tasks, we keep the number of tasks relatively small to ensure a reasonable human evaluation workload for fine-grained analysis.

\subsection{Metrics with Better Human Alignment}
Previous dynamic evaluation typically employs a binary success rate or a completion rate against a ``ground truth'' trajectory to evaluate the level of task completeness. However, real-world tasks often do not have a binary success flag or a single ground truth action sequence. For example, some tasks, such as ``Plan a one-day itinerary for Palo Alto,'' may involve exploration and information aggregation, where multiple reasonable solutions might exist. 
Thus, we seek to measure \textit{human satisfaction} rather than exact matches with a ground truth trajectory. For each task, we first manually write a list of rubrics (an example shown in Figure~\ref{fig:rubric_single_task}), containing both milestone steps (e.g., ``Opened Tripadvisor'') and satisfaction criteria (e.g., ``Viewed multiple attractions''). 
We then define the \textbf{Satisfaction Score (SS)} as the number of fulfilled rubrics divided by the total number of rubrics, as judged by a human evaluator.

We also include \textbf{Action Accuracy (AA)} and \textbf{Reflection Accuracy (RA)} as metrics to evaluate action-level performance. These metrics are also assessed by humans through a review of recorded screenshots and action histories. Finally, we include a \textbf{Termination Error (TE)} rate to reflect the agent’s robustness and error recovery capability. There are five ways an agent can exit from performing a task: (1) self-reported success: the agent decides to stop on its own; (2) reaching the maximum number of iterations: we set the maximum iteration count to 40 to prevent infinite loops; (3) reaching the maximum number of consecutive errors: if the agent has an action reflector and it identifies 3 consecutive errors, the agent is exited; (4) reaching the maximum number of repeated actions: if the agent performs the exact same action (excluding \texttt{Swipe} and \texttt{Back}) more than 3 consecutive times; (5) any other errors, such as errors when parsing the raw response into a valid action. 
If a task exits in one of the ways described in 2–5, it is marked as having a termination error. The TE rate is computed as the ratio of tasks with termination errors to all tasks.

\subsection{Evaluating Self-Evolving Mobile Agents} 
To the best of our knowledge, this is the first work exploring evaluation in cross-task evolution settings.
We consider two variants of \ours: with and without the self-evolution module.
When self-evolution module is enabled---referred to as \textit{\ours + Evo}---the agent performs sequentially across tasks within each scenario from the \dataset benchmark. The five tasks in a scenario share a persistent long-term memory. At the end of the $k$-th task, the Experience Reflectors are prompted to update the long-term memory based on the interaction history of the current task as well as the queries for the remaining $5-k$ tasks. This mimics the implicit requirement for an evolving agent to plan ahead, storing relevant knowledge for future interactions. 
In this setting, tasks performed later in the sequence benefit from a greater accumulation of Tips and Shortcuts, enabling us to analyze the progressive impact of self-evolution over time (detailed in Figure~\ref{fig:evo_thru_time}).

\begin{table*}[t]
\caption{Comparison with state-of-the-art models on the \dataset benchmark, using GPT-4o as the backbone. \ours outperforms previous SOTA models by a significant margin across all metrics, demonstrating superior long-term planning, decision accuracy, and error recovery. Enabling self-evolution (\textit{\ours + Evo}) further enhances performance. Reflection Accuracy for AppAgent and Mobile-Agent-v1 are omitted since they do not have action reflectors.
}
\label{tab:main_result_gpt4o}
\vspace{-10pt}
\begin{center}
\begin{small}
\resizebox{\textwidth}{!}
{
\centering
\setlength{\tabcolsep}{4pt} 
\begin{tabular}{lc|ccccc}
\toprule
Model & 
Type &
\begin{tabular}[c]{@{}c@{}} Satisfaction Score \\(\%)  $\uparrow$ \end{tabular} & 
\begin{tabular}[c]{@{}c@{}} Action Accuracy \\(\%)  $\uparrow$ \end{tabular} & 
\begin{tabular}[c]{@{}c@{}} Reflection Accuracy \\(\%) $\uparrow$ \end{tabular} & 
\begin{tabular}[c]{@{}c@{}} Termination Error \\(\%) $\downarrow$ \end{tabular} \\
\midrule
AppAgent~\cite{yang2023appagent}       & Single-Agent & 25.2 & 60.7  & - & 96.0 \\
Mobile-Agent-v1~\cite{wang2024mobile} & Single-Agent & 45.5 & 69.8  & - & 68.0 \\
Mobile-Agent-v2~\cite{wang2024mobile2} & Multi-Agent & 53.0 & 73.2  & 96.7 & 52.0 \\
\midrule
\ours & Multi-Agent & 75.1 & 85.9  & 97.4 & 32.0 \\
\ours + Evo         & Multi-Agent & \textbf{86.9} & \textbf{90.4} & \textbf{97.8} & \textbf{12.0} \\
\bottomrule
\end{tabular}
}
\end{small}
\end{center}
\vskip -0.15in
\end{table*}

\begin{table*}[t]
\caption{Results on different large multimodal model backbones, including GPT-4o, Gemini, and Claude. The metrics \textit{SS}, \textit{AA}, \textit{RA}, and \textit{TE} represent Satisfaction Score, Action Accuracy, Reflection Accuracy, and Termination Error, respectively, expressed as percentages.}
\label{tab:backbone_comparison}
\begin{center}
\begin{small}
{
\centering
\setlength{\tabcolsep}{4pt} 
\begin{tabular}{l|cccc|cccc|cccc}
\toprule
\multirow{2}{*}{Model} & 
\multicolumn{4}{c|}{Gemini-1.5-pro} &
\multicolumn{4}{c|}{Claude-3.5-Sonnet} &
\multicolumn{4}{c}{GPT-4o}\\
& 
SS$\uparrow$ & AA$\uparrow$ & RA$\uparrow$ & TE$\downarrow$ &
SS$\uparrow$ & AA$\uparrow$ & RA$\uparrow$ & TE$\downarrow$ &
SS$\uparrow$ & AA$\uparrow$ & RA$\uparrow$ & TE$\downarrow$
\\
\midrule
Mobile-Agent-v2~\cite{wang2024mobile2} & 50.8 & 63.4 & 83.9 & 64.0  & 70.9 & 76.4 & 96.9 & 32.0 &  53.0 & 73.2 & 96.7 & 52.0 \\
\midrule
\ours           & 70.9 & 74.3 & \textbf{91.3} & \textbf{48.0}  & 75.5 & 91.1 & 99.1 & \textbf{12.0} &  75.1 & 85.9 & 97.4 & 32.0 \\
\ours + Evo     & \textbf{71.2} & \textbf{77.4} & 89.6 & \textbf{48.0}  & \textbf{83.0} & \textbf{91.4} & \textbf{99.7} & \textbf{12.0} &  \textbf{86.9} & \textbf{90.4} & \textbf{97.8} & \textbf{12.0} \\
\bottomrule
\end{tabular}
}
\end{small}
\end{center}
\vskip -0.1in
\end{table*}

\subsection{Models}
\paragraph{Baselines.} We compare against a wide range of open-sourced mobile agent frameworks, including AppAgent~\cite{yang2023appagent}, Mobile-Agent-v1~\cite{wang2024mobile}, and Mobile-Agent-v2~\cite{wang2024mobile2}. To maximize an apple-to-apple comparison with Mobile-Agent-v2, which is the previous state-of-the-art, we apply an identical atomic operation space, perception model, and initial Tips to Mobile-Agent-v2 as \ours{}. AppAgent originally requires an additional exploration phase, which does not fit our setting; thus, we add the initial Tips as additional knowledge.

\vspace{-5pt}
\paragraph{Backbones.} We explore using various large multimodal models (LMM) as backbones for the reasoning agents, including GPT-4o~\cite{openai2024gpt4o}\footnote{GPT-4o version: gpt-4o-2024-11-20}, Claude-3.5-Sonnet~\cite{claude35}\footnote{Claude-3.5 version: claude-3-5-sonnet-20241022}, and Gemini-1.5-pro~\cite{team2024gemini15}\footnote{Gemini-1.5 version: gemini-1.5-pro-latest (Dec 2024)}. Unless otherwise specified, the default backbone for all models is GPT-4o.

\paragraph{Perceptor Implementation in \ours.} We closely follow Mobile-Agent-v2~\cite{wang2024mobile2} to implement the Perceptor with slight modifications. We use DBNet\footnote{\url{https://modelscope.cn/models/iic/cv_resnet18_ocr-detection-db-line-level_damo}}\cite{liao2020real_db_net} and ConvNextViT-document\footnote{\url{https://modelscope.cn/models/iic/cv_convnextTiny_ocr-recognition-document_damo}} from ModelScope for OCR detection and recognition respectively.
We use GroundingDINO~\cite{grounding_dino} for icon grounding and Qwen-VL-Plus~\cite{bai2023qwen} for generating captions for each cropped icon.
\section{Results}

\subsection{Evaluation on Performance}
\label{subset:result_main}

\paragraph{Comparison with state-of-the-art.} Table~\ref{tab:main_result_gpt4o} presents the results on \dataset using an identical GPT-4o backbone for all baselines and \ours.
\ours outperforms the previous multi-agent state-of-the-art (SOTA) model~\cite{wang2024mobile2} by \textbf{22.1\%} in the Satisfaction Score. This comparison particularly highlights the effectiveness of the hierarchy in our multi-agent framework.
Our approach also demonstrates superior robustness and error recovery capabilities, as indicated by a significantly lower termination error rate.
Moreover, enabling self-evolution further enhances performance, leading to an improvement of \textbf{33.9\%} against the previous SOTA, underscoring the benefit of learning from experience. 
In \S\ref{subsec:analysis_evo}, we provide further analysis of the impact of the evolution module.

\paragraph{Varying reasoning backbones.}
Table~\ref{tab:backbone_comparison} shows the comparison with previous SOTA~\cite{wang2024mobile2} using various backbone LMMs. We observe consistent improvements on all recent LMMs, including GPT-4o, Claude-3.5-Sonnet, and Gemini-1.5-pro, with average absolute gains of \textbf{22.1\%} and \textbf{15.6\%} with and without evolution, respectively. Additionally, the benefits of self-evolution appear to be more pronounced in stronger backbones, such as GPT-4o and Claude.


\begin{figure*}[ht]
\vskip 0.2in
\begin{center}
\subfigure[Single Task (task id: 5\_palo\_alto\_tour)]{
    \includegraphics[width=0.56\textwidth]{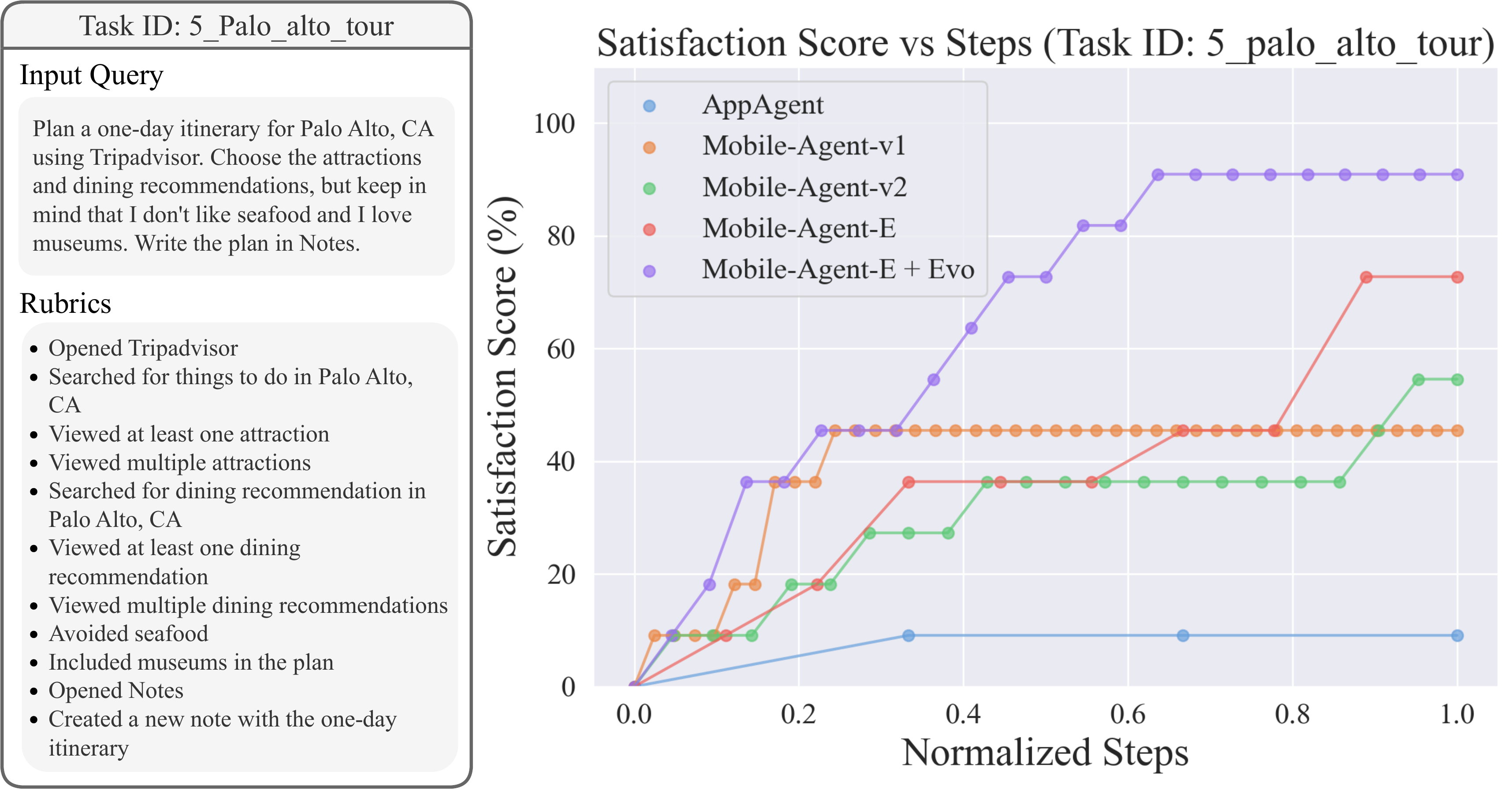}
    \label{fig:rubric_single_task}
}
\subfigure[All Tasks]{
    \includegraphics[width=0.40\textwidth]{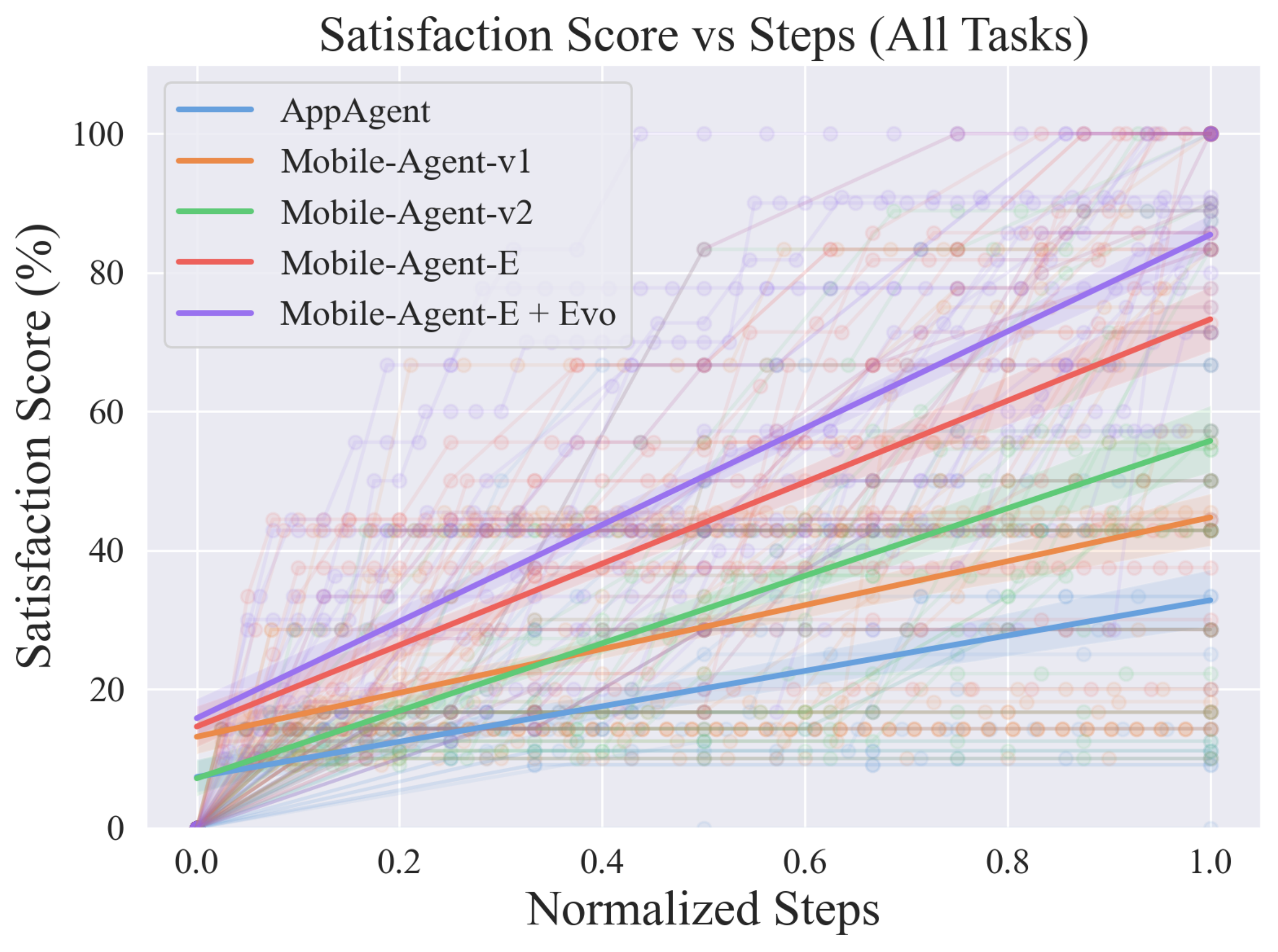}
    \label{fig:rubric_all_tasks}
}
\vspace{-10pt}
\caption{\small{Satisfaction Score vs. Steps (SSS) curve for (a) a single task and (b) all tasks. In (a), we also provide a concrete example of the human-written rubrics for the task, which are used to compute the Satisfaction Score during human evaluation. In (b), we additionally include a linear regression line for each model; a steeper and higher line indicates better efficiency for completing the task.}}
\label{fig:rubric_vs_steps_combined}
\end{center}
\vskip -0.2in
\end{figure*}

\vspace{-5pt}
\subsection{Evaluation on Efficiency}
\label{subsec:efficiency_results}
Evaluating the efficiency of mobile agents on complex, potentially open-ended tasks is not straightforward. Merely counting the number of steps is not optimal, as many tasks require exploration. A smaller number of steps reflects a quick exit but may result in insufficient exploration.
Intuitively, if an agent achieves higher satisfaction, i.e., fulfills more rubrics, in a smaller number of steps, it is considered more efficient. Thus, we introduce the \textbf{Satisfaction Score vs Steps (SSS) curve} to compare and visualize the efficiency of different agents.
To plot the SSS curve, we manually examine the recorded trajectories and track the satisfaction of rubrics after each step. We then plot a poly-line with the step number as the x-axis and the Satisfaction Score as the y-axis. To enable visualization of trajectories with different lengths on the same graph, we normalize the steps to the range [0, 1].
The y-axis of the rightmost point indicates the final satisfaction score. Intuitively,\textbf{ a steeper and higher SSS curve indicates better efficiency and completeness}. As shown in Figure~\ref{fig:rubric_vs_steps_combined}, we observe that \ours not only achieves better final performance but also fulfills rubrics faster.

\begin{figure}[t]
\begin{center}
\centerline{\includegraphics[width=0.9\columnwidth]{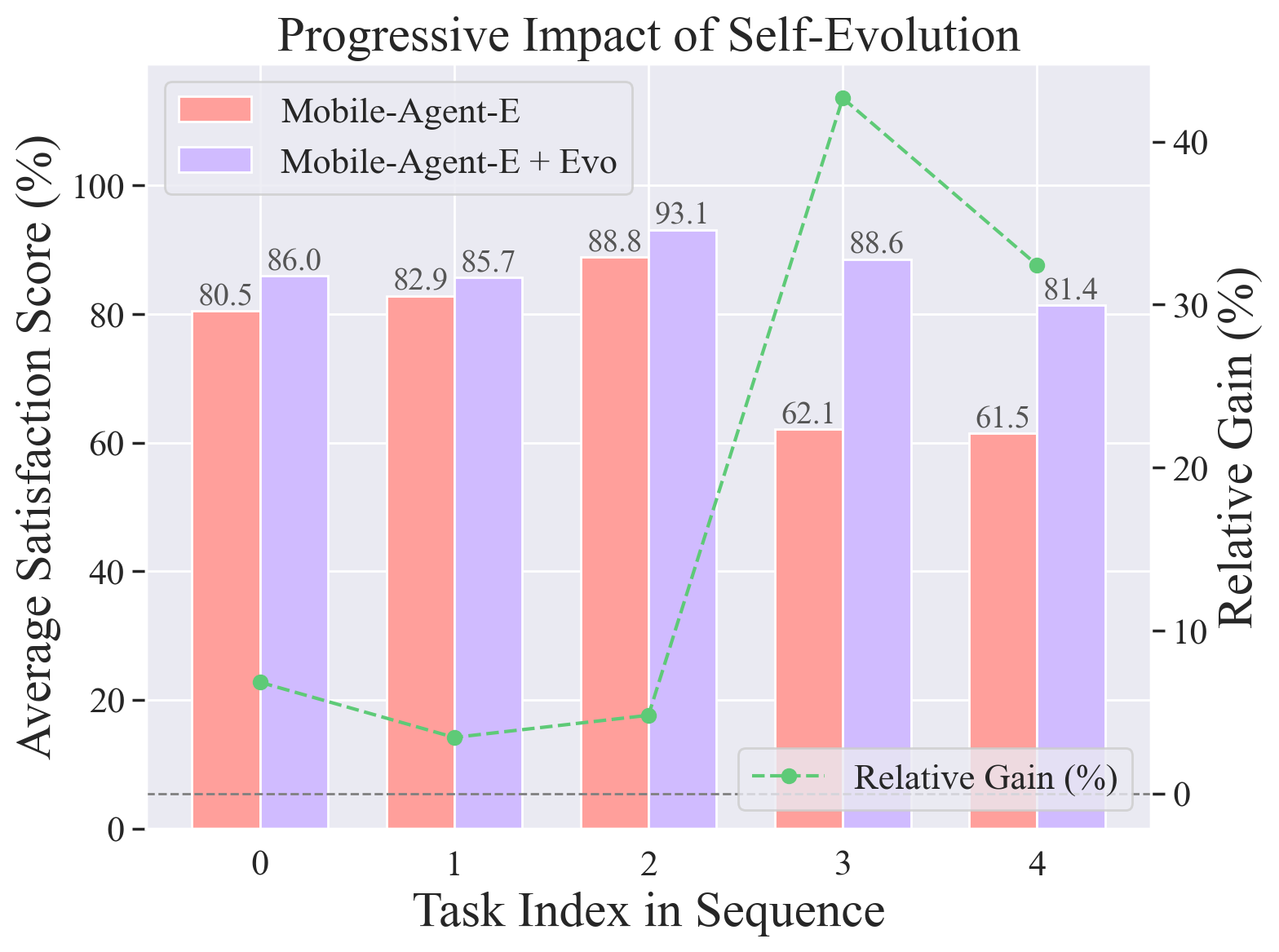}}
\vspace{-10pt}
\caption{Progressive impact of self-evolution over time. The task index represents the order in which a task is performed in the evolution setting. The results demonstrate that tasks performed later in the sequence show more significant improvements, highlighting the increased benefits from additional iterations of self-evolution.}
\label{fig:evo_thru_time}
\end{center}
\vskip -0.4in
\end{figure}

\vspace{-5pt}
\subsection{Further Analysis}
\label{subsec:analysis_evo}

\paragraph{Progressive impact of self-evolution over time.}
The ideal behavior of self-evolution is to progressively bring more benefits to the agent as knowledge accumulates. To investigate this, we group the results of the tasks by their ordering index in each scenario and compare the performance with and without enabling the evolution module. In Figure~\ref{fig:evo_thru_time}, the x-axis reflects the task index in the sequence it is performed, with later tasks having access to Tips and Shortcuts that are updated through more tasks.
We observe a generally increasing trend indicating that the gain tends to be more significant in later tasks, demonstrating that the self-evolution module is capable of continuously improving the agent as it experiences more tasks. The gain is not strictly monotonically increasing, as expected, since the difficulty of tasks at different indices varies.

\begin{table}[t]
\caption{
Analysis of computational overhead and Shortcut usage. In the inference speed table, the \textit{reasoning only} section accounts for time spent solely on reasoning agents, while \textit{perception + reasoning} includes the runtime of the Perceptor \textbf{on CPU}. Shortcut usage statistics are calculated as the ratio of Shortcuts used to the total number of actions performed by the Operator. The use of Shortcuts significantly accelerates inference, achieving comparable times to previous, simpler frameworks.
}
\label{tab:speed}
\vspace{-5pt}
\begin{center}
\begin{small}
\resizebox{0.49\textwidth}{!}
{
\centering
\setlength{\tabcolsep}{3pt} 
\begin{tabular}{l|ccc|ccc}
\toprule
\multicolumn{7}{c}{Inference Speed (Seconds per operation)}\\
\toprule
\multirow{2}{*}{Model} &
\multicolumn{3}{c|}{\textit{Reasoning Only}} &
\multicolumn{3}{c}{\textit{Perception + Reasoning}}\\
 &
\begin{tabular}[c]{@{}c@{}} Gemini \end{tabular} &
\begin{tabular}[c]{@{}c@{}} Claude \end{tabular} &
\begin{tabular}[c]{@{}c@{}} GPT \end{tabular} &
\begin{tabular}[c]{@{}c@{}} Gemini \end{tabular} &
\begin{tabular}[c]{@{}c@{}} Claude \end{tabular} &
\begin{tabular}[c]{@{}c@{}} GPT \end{tabular}
\\
\midrule
Mobile-Agent-v2  & 9.8 & 21.4 & 12.3 & 25.6 & 38.4 & 43.5 \\
\ours  & 16.5 & 25.5 & 17.4 & 30.8 & 41.0 & 30.1 \\
\ours + Evo  & 12.9 & 24.8 & 14.9 & 27.2 & 39.6 & 27.4\\
\bottomrule
\end{tabular}
}

\vskip 0.05in
\resizebox{0.37\textwidth}{!}
{
\begin{tabular}{l|ccc}
\toprule
\multicolumn{4}{c}{Shortcut Usage Percentage (\%)}\\
\toprule
Model &
Gemini &
Claude &
GPT \\
\midrule
\ours & 11.9 & 12.8 & 12.4 \\
\ours + Evo & 14.8 & 13.2 & 14.4\\
\bottomrule
\end{tabular}
}

\end{small}
\end{center}
\vskip -0.3in
\end{table}

\paragraph{Shortcut reduces computational overhead.}
The hierarchical multi-agent architecture in \ours significantly improves performance on complex tasks but inevitably increases computational complexity. However, we found that the use of Shortcuts largely mitigates this overhead, enabling \ours to achieve a speed comparable to that of previous models.
In Table~\ref{tab:speed}, we report the seconds per operation averaged across all tasks as well as the usage of Shortcuts. 
We observe a positive correlation between using more Shortcuts and faster inference speed.
This is because a Shortcut enables the execution of multiple operations within a single decision-making iteration. For example, using the \texttt{Tap\_Type\_and\_Enter} Shortcut to perform a search subroutine saves two iterations of perception and reasoning compared to using three atomic actions: \texttt{Tap}, \texttt{Type}, and \texttt{Enter}.

\begin{figure*}[t]
\vskip 0.2in
\begin{center}
\centerline{\includegraphics[width=0.95\textwidth]{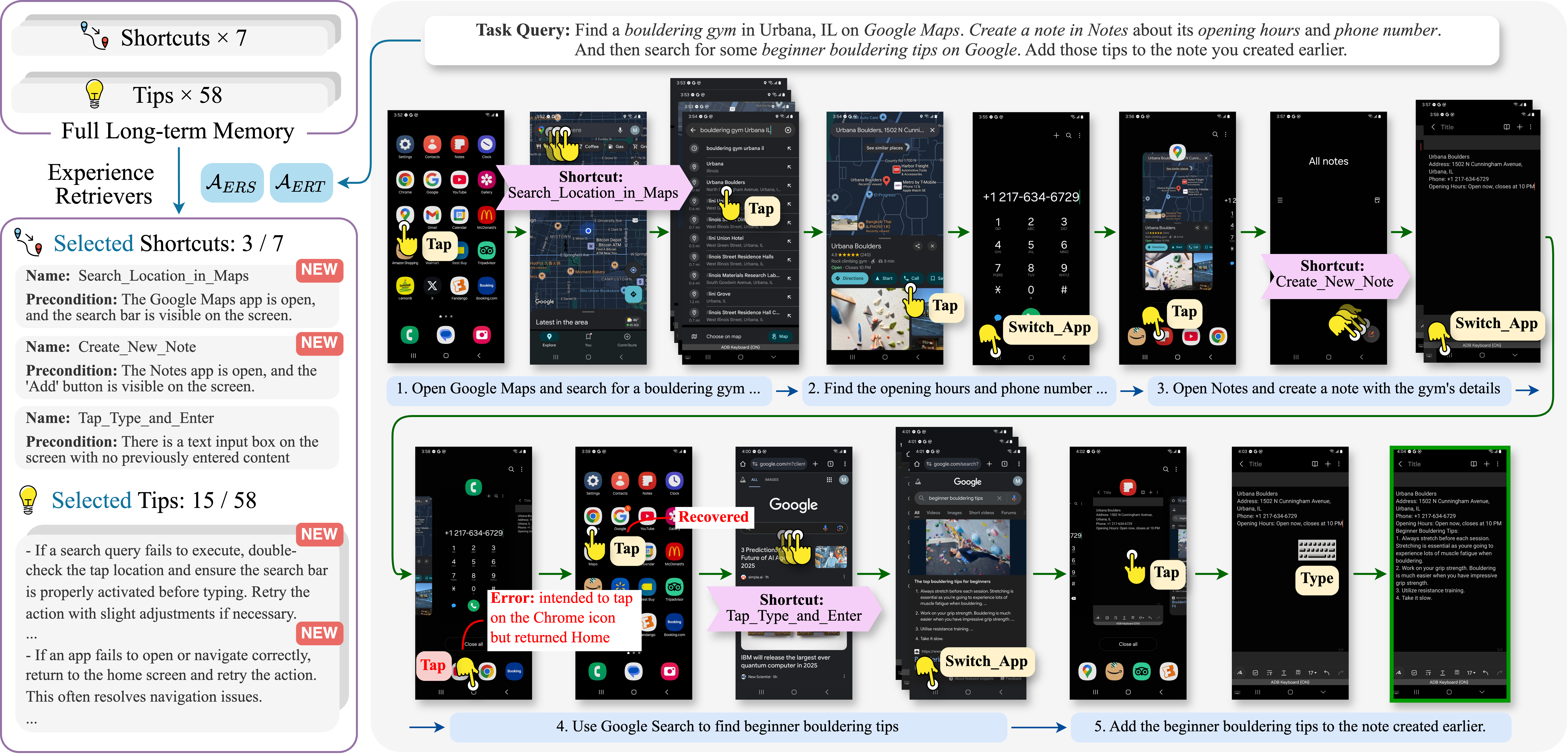}}
\vspace{-5pt}
\caption{Case study example where relevant Shortcuts and Tips are automatically retrieved from the previously evolved long-term memory and subsequently leveraged to complete an unseen, challenging task. The action trajectory also includes an example where the agent recovers from an error.
}
\label{fig:case_study}
\end{center}
\vskip -0.3in
\end{figure*}

\begin{table}[t]
\caption{To investigate the unique impact of Tips, we compute the Satisfaction Score on a subset of instances where no newly generated Shortcuts are used in the trajectory. The results show distinctive benefits from the evolved Tips.}
\label{tab:tips_only_impact}
\begin{center}
\begin{small}
\resizebox{0.48\textwidth}{!}
{
\centering
\setlength{\tabcolsep}{4pt} 
\begin{tabular}{lccc}
\toprule
 & 
\begin{tabular}[c]{@{}c@{}} Gemini \end{tabular} &
\begin{tabular}[c]{@{}c@{}} Claude \end{tabular} &
\begin{tabular}[c]{@{}c@{}} GPT-4o \end{tabular}\\
\midrule
\ours & 69.0 & 75.6 & 79.7  \\
\ours + evolved Tips & \textbf{72.6} & \textbf{85.2} & \textbf{87.5} \\
\bottomrule
\end{tabular}
}
\end{small}
\end{center}
\vskip -0.15in
\end{table}

\paragraph{Unique impact from Tips.}
While the impact from Shortcuts is directly visible in the action history, it is less obvious whether the evolved Tips bring distinctive benefits. To visualize this, we filter out task instances where the same set of unique Shortcuts is used or where only atomic actions are employed, and compare the Satisfaction Score with or without the evolution. Table~\ref{tab:tips_only_impact} shows that Tips alone serve as an important aspect of self-evolution.

\subsection{Case Study: A Closed-Loop Self-Evolving Agent}
In real-world mobile usage, after running the agent on a large number of tasks in various scenarios, the accumulated Tips and Shortcuts may grow to an amount where it is no longer feasible to include all of them in the decision-making context.
Thus, in this case study, we aim to explore closing the self-evolution loop by introducing two additional \textbf{Experience Retriever} agents for Tips $\mathcal{A}_{ERT}$ and Shortcuts $\mathcal{A}_{ERS}$.
We consider a new task in an unknown scenario, as shown in Figure~\ref{fig:case_study}.
First, we provide all the updated Tips and Shortcuts---after running \ours on all 5 scenarios (a total of 25 tasks) in \dataset---to the Experience Retrievers. With GPT-4o as the backbone, the updated long-term memory contains a total of 7 unique Shortcuts and 58 Tips, among which 6 Shortcuts and 55 Tips are newly proposed by \ours during experience reflection.
Then, the Experience Retrievers are prompted to select only the relevant Tips and Shortcuts for the current task.
The qualitative example in Figure~\ref{fig:case_study} shows that \ours effectively retrieves and leverages highly relevant Shortcuts and Tips to successfully complete a challenging unseen task.
The full list of Tips and Shortcuts after evolution can be found in Appendices~\ref{app:full_list_of_evolved_tips} and \ref{app:full_list_of_evolved_shortcuts}.

\section{Related Work}
\label{sec:related_work}

\subsection{GUI Agents}
The advancement of large multimodal models (LMM) has introduced a new area of agentic research focused on LMM-based GUI agents~\cite{wang2024guisurvey}. The goal is to develop AI assistants capable of performing tasks in various GUI environments, such as Web~\cite{deng2023mindweb, zheng2024seeact, he2024webvoyager, yoran2024assistantbenchwebagentssolve, reddy2024infogent}, PC~\cite{hong2023cogagent, ufo, liu2024visualagentbench, xie2024osworld, cradle}, and mobile devices~\cite{wang2024mobile, yang2023appagent, li2024appagentv2, wang2024mobile2, liu2024autoglm}.
In the mobile environment, one line of research focuses on improving the perception and reasoning abilities of a single agent through tool usage~\cite{wang2024mobile} and an additional exploration phase~\cite{yang2023appagent, li2024appagentv2}.
Recent studies~\cite{rawles2024androidworld, wang2024mobile2} show significant promise by incorporating multiple agents for decision-making and reflection. However, current multi-agent frameworks still face challenges such as short-sighted planning and poor error recovery.
Specifically, the ``planning'' module in Mobile-Agent-v2~\cite{wang2024mobile2} functions primarily as a progress tracker, while the ``decision-making'' module continues to handle both high-level planning (e.g., ``what to do next'') and low-level action decisions (e.g., ``where to tap'').
A key difference in \ours is the introduction of a \textit{hierarchy} among the agents, enabling more effective long-horizon planning and improved low-level action accuracy.

\subsection{Self-Evolution in Foundation Models}
Investigating how to make large language models and multimodal models self-improve has long been an active area of research~\cite{tao2024self_evolve_survey}. One line of work focuses on enhancing the base abilities of foundation models, such as improving reasoning and reducing knowledge hallucination. This includes approaches like iterative refinement~\cite{madaan2024selfrefine}, self-reflection~\cite{shinn2024reflexion}, self-training~\cite{huang2022llm_self_improve}, self-improvement~\cite{selfimprovement2024a}, and multi-persona collaboration~\cite{wang2023unleashing}.
Another line of work explores improving task-solving with foundation models through tool learning and tool creation~\cite{cai2023large_tool_maker, qian2023creator, yuan2023craft}.
In the context of GUI agents, self-evolution is less studied. The skill curation mechanism in Cradle~\cite{cradle} shows initial promise in the PC environment; however, no previous work has systematically explored self-evolution in mobile environments. In this work, we demonstrate the importance of self-evolution in both Tips and Shortcuts. Notably, unlike the ``skills’’ in Cradle, which are directly added to the atomic operation space, we explicitly define \textit{preconditions} for our Shortcuts, as this is critical for decision-making across multiple apps and varying layouts.

\section{Conclusion and Future Work}
\label{sec:conclusion}
We introduce \ours, a novel mobile assistant featuring a hierarchical multi-agent framework and a self-evolution module that significantly enhances long-term planning, error recovery, and efficiency, excelling in a wide variety of complex real-world tasks. Remaining limitations include the incorrect usage of Shortcuts with invalid preconditions and erroneous agent-generated Shortcuts, with detailed examples provided in Appendix~\ref{app:remaining_limitations}. Future work will focus on developing improved strategies for generating, invoking, and revising Shortcuts, enhancing personalization to better adapt to individual user needs, and strengthening safety precautions to enable more effective human-agent collaboration.




\section*{Impact Statement}
This paper aims to advance the field of LMM-based agents by developing a hierarchical multi-agent framework and benchmark to improve the usability and efficiency of smartphones in complex, multi-step tasks. While the primary goal is to enhance human-device interaction, the proposed system has the potential for broader societal benefits, particularly in improving accessibility for individuals with disabilities or limited mobility. By enabling more intuitive and automated task management on mobile devices, this framework can assist users with physical impairments, cognitive challenges, or conditions that make precise interactions with touchscreens difficult.

While the primary aim is to enhance mobile task efficiency and user accessibility, the development of mobile agents capable of autonomous decision-making introduces potential risks. For example, unauthorized or unintended actions by the agent, such as the misuse of sensitive information including credit card details or private data, could result in serious consequences for users.
These risks emphasize the critical need for robust safeguards, error recovery mechanisms, and fail-safe systems to ensure that the agent’s actions consistently align with user intentions.

We are actively pursuing future work that focuses on designing and integrating robust privacy and safety mechanisms.
These include explicit user consent workflows for sensitive operations, encryption protocols to protect user data during processing and storage, and automated systems to flag potentially harmful or unauthorized actions.
These advancements will be crucial for maximizing the societal benefits of these systems, minimizing potential risks, and building user trust in autonomous mobile agents.

\newpage

\bibliography{custom}
\bibliographystyle{icml2025}

\newpage
\appendix
\onecolumn

\begin{figure*}[t]
\vskip 0.2in
\begin{center}
\centerline{\includegraphics[width=0.98\textwidth]{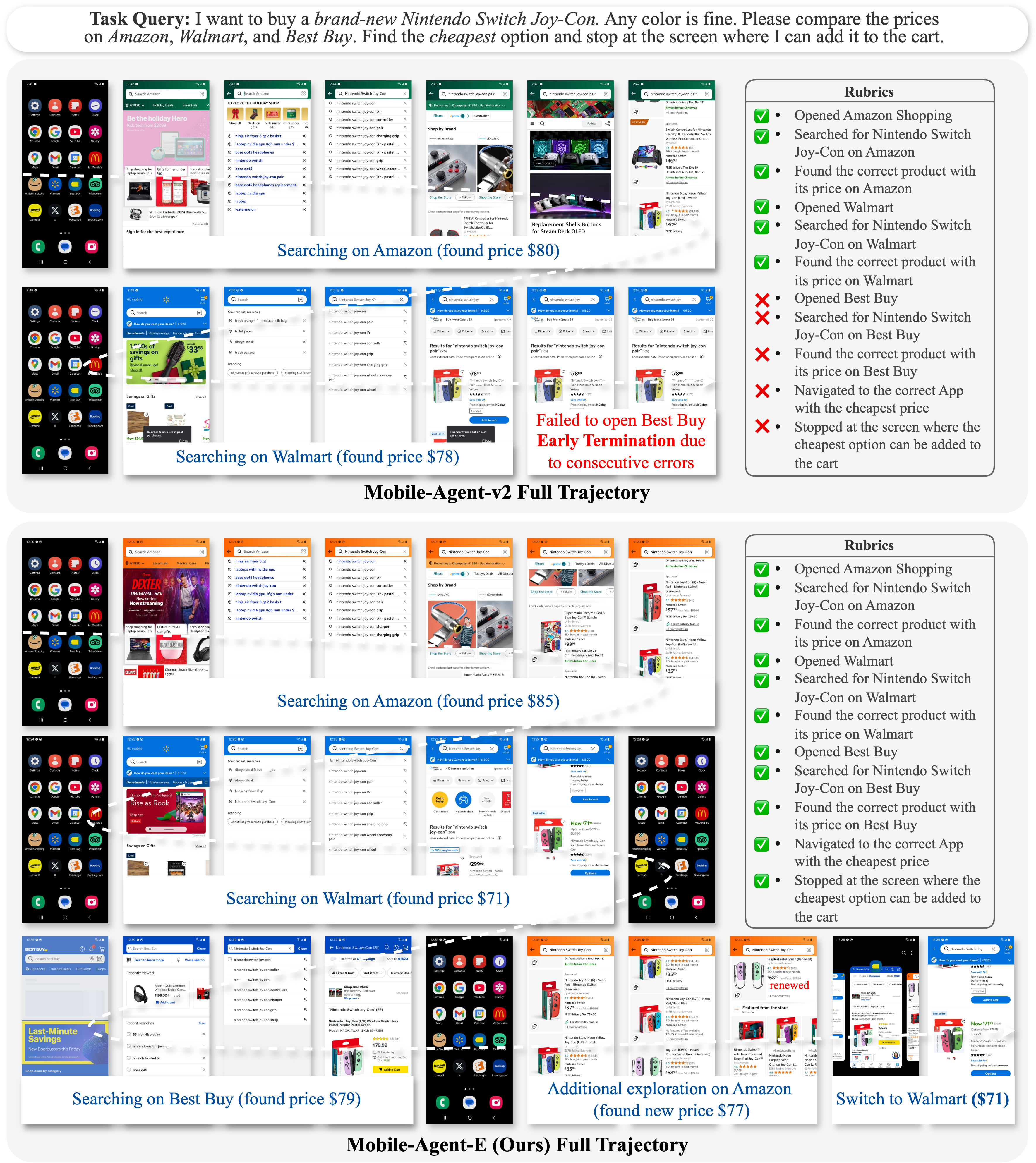}}
\caption{Full trajectory comparison between the previous state-of-the-art, Mobile-Agent-v2~\cite{wang2024mobile2}, and \ours.}
\label{fig:full_trajectory_comparison}
\end{center}
\vskip -0.2in
\end{figure*}

\section{Full Trajectory Comparison Example with Previous SOTA}
\label{app:full_traj_example}

Figure~\ref{fig:full_trajectory_comparison} presents the full trajectory of the task shown in Figure~\ref{fig:teaser}, comparing the previous state-of-the-art, Mobile-Agent-v2~\cite{wang2024mobile2}, and our proposed \ours. Mobile-Agent-v2 suffers from early termination after interacting with two Apps, whereas \ours fulfills all rubrics and stops at the App offering the best deal.

\begin{figure*}[t]
\vskip 0.2in
\begin{center}
\centerline{\includegraphics[width=0.95\textwidth]{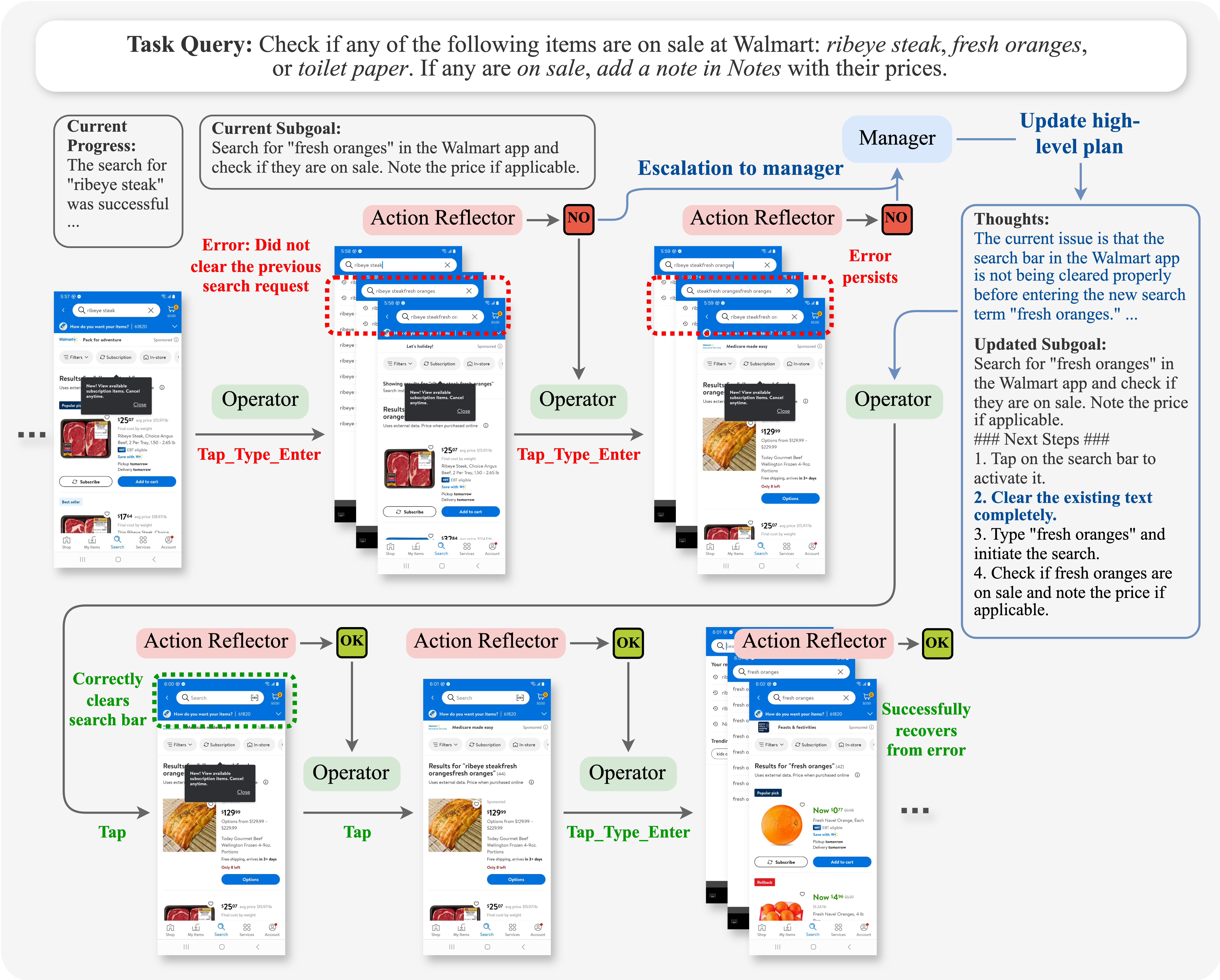}}
\caption{Error recovery with escalation. The task requires the agent to search for three different items on Walmart and note their sales information. At the step shown in the figure, the agent has already searched for ribeye steak and intends to search for fresh oranges next. However, the Operator erroneously calls the Shortcut that inputs text into the search bar and performs a search without clearing the previously entered text. Although the Action Reflector raises an error, the subgoal remains unchanged, and the Operator fails to rectify the error on the second attempt. After observing two consecutive errors, the error is escalated to the Manager, which correctly identifies the problem and revises the subgoal with detailed, decomposed steps to address the error. This helps the Operator correctly recover from the previous error by first tapping the ``$\times$'' icon to clear the previous search query.}
\label{fig:error_escalation}
\end{center}
\vskip -0.2in
\end{figure*}

\section{Error Recovery with Escalation to Manager}
\label{app:error_escalation}
Figure~\ref{fig:error_escalation} illustrates how the error escalation mechanism in \ours enhances error recovery ability. A detailed description of the example is provided in the caption.

\begin{figure*}[t]
\vskip 0.2in
\begin{center}
\centerline{\includegraphics[width=0.95\textwidth]{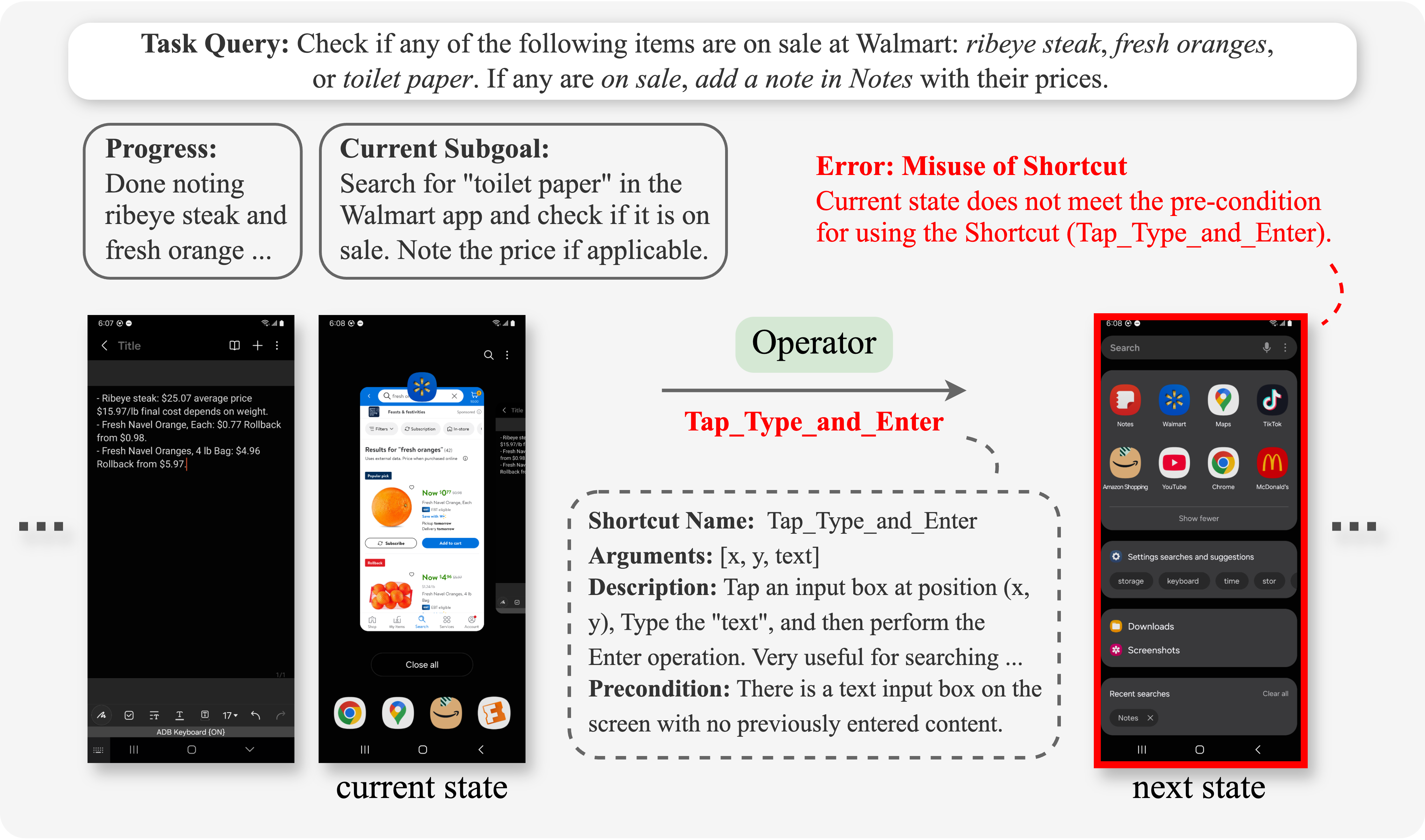}}
\vspace{-5pt}
\caption{Example of misuse of Shortcuts in an invalid state. At the current step, as shown in the figure, the agent intended to switch back to Walmart to search for the final item requested by the user. While it correctly performs the ``Switch\_App'' operation, it then calls a Shortcut for searching without realizing that it is not yet in the App where the search bar is available.}
\label{fig:error_in_action}
\end{center}
\vskip -0.2in
\end{figure*}

\section{Remaining Limitations}
\label{app:remaining_limitations}

\subsection{Misuse of Shortcuts due to Incorrect Perception of Phone State}
Although we explicitly require the Operator to verify the current phone state to ensure it fulfills the \textit{precondition} of a Shortcut before calling it, there are still cases where the model incorrectly perceives the state, resulting in the misuse of Shortcuts in an invalid state. Figure~\ref{fig:error_in_action} illustrates an example of such error. A detailed description of the example is provided in the caption. This type of error could potentially be mitigated by employing a dedicated agent for verifying preconditions or by enhancing the perception module to better understand phone states.

\begin{figure*}[t]
\vskip 0.2in
\begin{center}
\centerline{\includegraphics[width=0.95\textwidth]{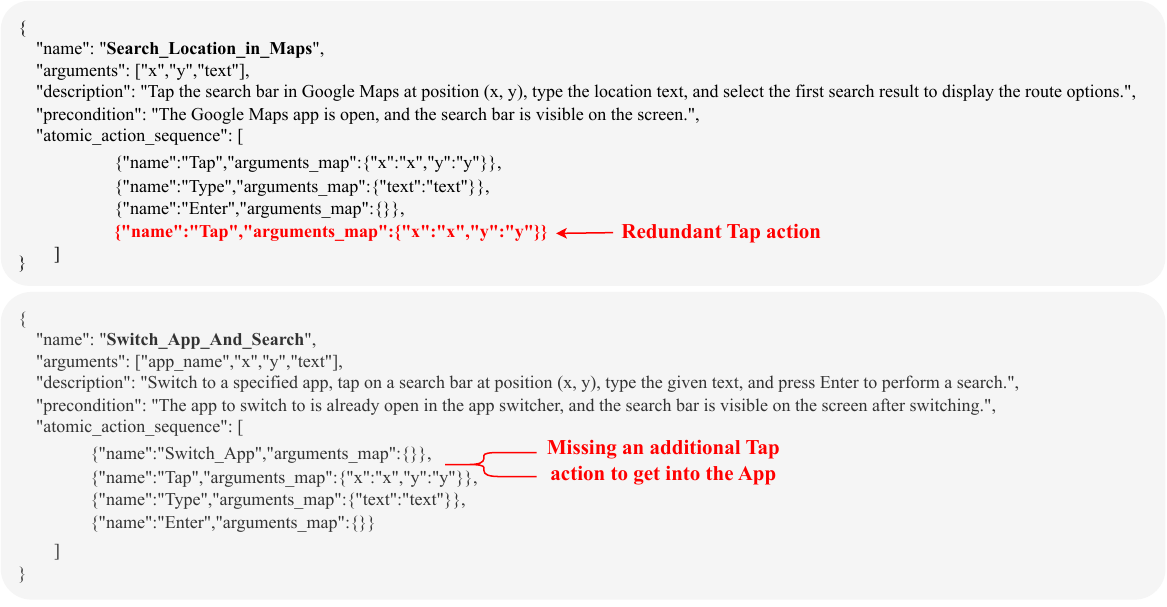}}
\caption{Example of imperfect (above) and erroneous (below) generated Shortcuts. The ``Search\_Location\_in\_Maps'' Shortcut includes an unnecessary Tap action in the operation sequence, while the ``Switch\_App\_And\_Search'' Shortcut omits a Tap action needed to first enter the desired App before performing the search.}
\label{fig:error_in_shortcut}
\end{center}
\vskip -0.2in
\end{figure*}

\subsection{Errors and Imperfections in Self-Evolved Shortcuts}
Although effective in most cases, we still observe errors and imperfections in the agent-generated Shortcuts during self-evolution. These issues can lead to propagated errors when an erroneous Shortcut is used in subsequent tasks. Figure~\ref{fig:error_in_shortcut} illustrates an example of such erroneous and imperfect Shortcuts. A detailed description of the example is provided in the caption. This highlights the need for future work on approaches to generate higher-quality Shortcuts and equipping the agent with the ability to reflect on and revise generated Shortcuts in subsequent tasks.

\section{All Tasks in \dataset Benchmark}
\label{app:all_task_our_benchmark}

\begin{table*}[ht]
\caption{All task queries in \dataset{}.}
\label{tab:all_benchmark_tasks}
\vskip 0.15in
\begin{center}
\scriptsize{
\setlength{\tabcolsep}{4pt} 
\begin{tabular}{p{1.8cm} p{2.8cm} p{1.5cm} p{9.7cm}}
\toprule
\textbf{Scenario} & 
\textbf{Task ID} & 
\textbf{APPs} &
\textbf{Input Query} \\
\midrule
\multirow{13}{*}{
\begin{tabular}[c]{@{}l@{}} Restaurant \\Recommendation \end{tabular}} 
& 1\_late\_night\_korean\_food & Maps & Find the best-rated late-night Korean restaurant in Champaign, IL that opens beyond 9pm on Google Maps. \\
\cmidrule(lr){2-4}
& 1\_nearest\_bakery & Maps & Get directions to the nearest Bakery that has a rating higher than 4.0 on Google Maps. Stop at the screen showing the route. \\
\cmidrule(lr){2-4}
& 1\_thai\_duck & Maps, Notes & Find the best-rated Thai restaurant in Urbana, IL that serves duck cuisine on Google Maps. Review customer comments and compile a summary of positive and negative feedback in Notes. \\
\cmidrule(lr){2-4}
& 1\_bakery\_birthday\_cake & Maps, Notes & Find me a Bakery that is within 10min drive near me and does birthday cakes on Google Maps. Find the phone number and create a new note in Notes for that. \\
\cmidrule(lr){2-4}
& 1\_chinese\_ohare & Maps, X, Notes & Find me a popular Chinese restaurant near Chicago O'Hare airport on Google Maps. Check X for recent posts about their signature dishes and write a summary in Notes. Then get directions to that restaurant on Google Maps. Stop at the screen showing the route. \\
\midrule
\multirow{13}{*}{
\begin{tabular}[c]{@{}l@{}} Information \\Researching \end{tabular}} 
& 2\_segment\_anything\_cited & Chrome & Find the most-cited paper that cites the paper 'Segment Anything' on Google Scholar. Stop at the screen showing the paper abstract. \\
\cmidrule(lr){2-4}
& 2\_llm\_agents\_survey & Chrome, Notes & Find at least three representative survey papers on LLM agents on Google Scholar, and add their titles to the Notes. \\
\cmidrule(lr){2-4}
& 2\_recipes\_chinese & Chrome, YouTube & I have some onions, beef, and potatoes in my refrigerator. Can you find me a Chinese-style recipe that uses all three ingredients and can be prepared in under an hour? And find me a video tutorial on YouTube for that. Stop at the screen displaying the video. \\
\cmidrule(lr){2-4}
& 2\_mcdonalds\_deals & McDonald's, Maps & Can you check the McDonald's APP to see if there are any Rewards or Deals including Spicy McCrispy. If so, help me add that to Mobile Order (Do not pay yet, I will do it myself). And then check the pickup location and get directions on Google Maps. Stop at the screen showing the route. \\
\cmidrule(lr){2-4}
& 2\_headphones\_reviews & Amazon, Notes & Find three detailed user reviews of the Bose QC45 headphones from Amazon. Summarize the general sentiment in the Notes. \\
\midrule
\multirow{13}{*}{
\begin{tabular}[c]{@{}l@{}} Online \\Shopping \end{tabular}} 
& 3\_oled\_tv & Best Buy & Find the best deal on a 55-inch 4K OLED TV at Best Buy. Stop at the screen displaying the best deal you find. \\
\cmidrule(lr){2-4}
& 3\_laptop\_nvidia\_gpu & Amazon Shopping & Find me a laptop on Amazon that is under \$1000 with an Nvidia GPU and more than 8GB RAM. \\
\cmidrule(lr){2-4}
& 3\_ninja\_air\_fryer & Amazon Shopping, Walmart & Compare the price of a Ninja air fryer 8 qt at Walmart and Amazon. Stop at the screen displaying the best deal you find. \\
\cmidrule(lr){2-4}
& 3\_walmart\_sale\_items & Walmart, Notes & Check if any of the following items are on sale at Walmart: ribeye steak, fresh oranges, or toilet paper. If any are on sale, add a note in Notes with their prices. \\
\cmidrule(lr){2-4}
& 3\_nintendo\_switch\_joy\_con & Amazon Shopping, Best Buy, Walmart & I want to buy a brand-new Nintendo Switch Joy-Con. Any color is fine. Please compare the prices on Amazon, Walmart, and Best Buy. Find the cheapest option and stop at the screen where I can add it to the cart. \\
\midrule
\multirow{13}{*}{
\begin{tabular}[c]{@{}l@{}} What's \\Trending \end{tabular}} 
& 4\_x\_black\_myth\_wukong & X, Notes & Find the top posts about the game 'Black Myth Wukong' on X and summarize the key highlights in Notes. \\
\cmidrule(lr){2-4}
& 4\_x\_trending\_news & X, Notes & Check the top 3 trending news on X. Read a few posts to figure out what's happening. And create a new Note to summarize your findings. \\
\cmidrule(lr){2-4}
& 4\_watercolor\_painting\_tutorial & Lemon8, Notes & I want to learn how to paint watercolor. Find me some content creators to follow on Lemon8 that has highly liked posts about watercolor painting tutorials. List their account names in Notes. \\
\cmidrule(lr){2-4}
& 4\_movie\_trending & Fandango, Notes & Check the top 5 trending movies on Fandango that are currently in theaters. Compare their ratings and create a note in Notes for the highest-rated one, including its name and showtimes. \\
\cmidrule(lr){2-4}
& 4\_horror\_movie\_reviews & Fandango, Lemon8, Notes & Find me the latest horror movie currently in theaters on Fandango. Check some reviews on Lemon8 about the movie and create a note in Notes with the general sentiment. \\
\midrule
\multirow{13}{*}{
\begin{tabular}[c]{@{}l@{}} Travel \\Planning \end{tabular}} 
& 5\_cheap\_flights\_newyork & Booking & Find the cheapest round-trip flight from Chicago to New York City in the next month on Booking. Stop at the screen showing the best deal. \\
\cmidrule(lr){2-4}
& 5\_things\_to\_do\_la & Tripadvisor, Notes & Suggest some interesting things to do in LA. Find the top 3 attractions on Tripadvisor. Save the list in Notes. \\
\cmidrule(lr){2-4}
& 5\_palo\_alto\_tour & Tripadvisor, Notes & Plan a one-day itinerary for Palo Alto, CA using Tripadvisor. Choose the attractions and dining recommendations, but keep in mind that I don't like seafood and I love museums. Write the plan in Notes. \\
\cmidrule(lr){2-4}
& 5\_local\_food\_chicago & Tripadvisor, Notes & Find a highly recommended local restaurant in Chicago on Tripadvisor. Check the reviews about must-try dishes and summarize in Notes. \\
\cmidrule(lr){2-4}
& 5\_hotel\_champaign & Booking, Maps & Help me find a hotel in Champaign, IL on Booking that is under \$200 for a queen bed. Make sure that the rating is higher than 7.0. Double-check on Google Maps to see if it is close to Green Street. Show me your final choice on Booking. \\
\bottomrule
\end{tabular}
}
\end{center}
\vskip -0.1in
\end{table*}

Table~\ref{tab:all_benchmark_tasks} presents the input queries, involved App types, and scenarios for all \dataset tasks.
The complete list of rubrics and human reference operation sequences is provided in the supplementary material.

\section{Atomic Operation Space}
\label{app:full_atomic_operation_space}
\begin{table}[ht]
\caption{Atomic operations space.}
\label{tab:atomic_operation_space}
\vskip 0.15in
\begin{center}
\begin{small}
{
\centering
\setlength{\tabcolsep}{4pt} 
\begin{tabular}{p{3.5cm} p{12cm}}
\toprule
\textbf{Operation} & 
\textbf{Description} \\
\midrule
$Open\_App(app\_name)$ & If the current screen is Home or App screen, you can use this action to open the app named ``app\_name'' on the visible on the current screen. \\
\midrule
$Tap(x, y)$ & Tap the position (x, y) in current screen.\\
\midrule
$Swipe(x_1, y_1, x_2, y_2)$& Swipe from position $(x_1, y_1)$ to position $(x_2, y_2)$. To swipe up or down to review more content, you can adjust the y-coordinate offset based on the desired scroll distance. For example, setting $x_1 = x_2 = 0.5 * width,\; y_1 = 0.5 * height$, and $y_2 = 0.1 * height$ will swipe upwards to review additional content below. To swipe left or right in the App switcher screen to choose between open apps, set the x-coordinate offset to at least $0.5 * width$.\\
\midrule
$Type(text)$& Type the "text" in an input box.\\
\midrule
$Enter()$& Press the Enter key after typing (useful for searching). \\
\midrule
$Switch\_App()$ & Show the App switcher for switching between opened apps.\\
\midrule
$Back()$& Return to the previous state.\\
\midrule
$Home()$& Return to home page.\\
\midrule
$Wait()$& Wait for 10 seconds to give more time for a page loading.\\
\bottomrule
\end{tabular}
}
\end{small}
\end{center}
\vskip -0.1in
\end{table}

Table~\ref{tab:atomic_operation_space} presents all atomic operations considered in \ours.

\begin{figure*}[t]
\vskip 0.2in
\begin{center}
\centerline{\includegraphics[width=0.98\textwidth]{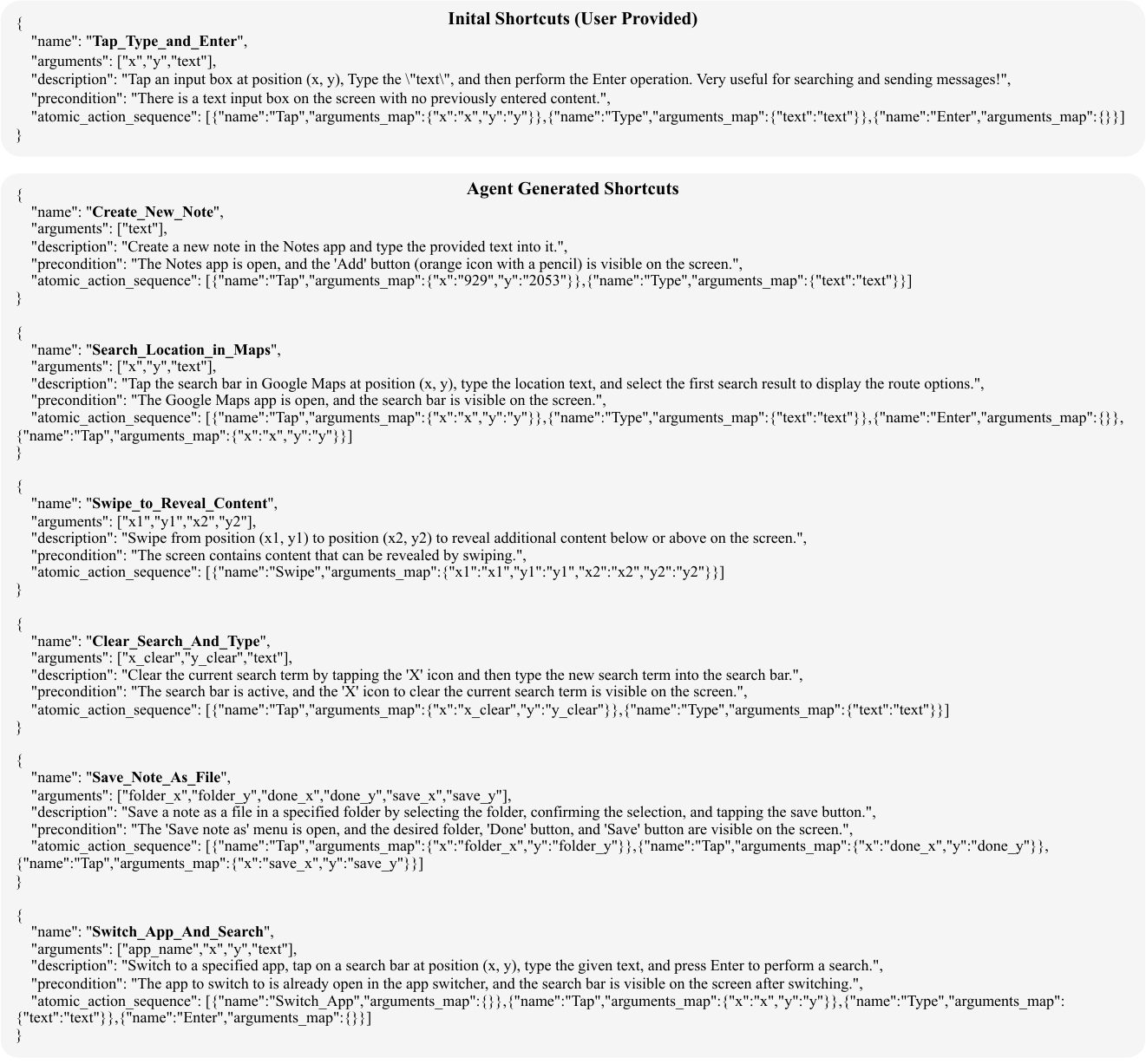}}
\caption{Full list of Shortcuts generated by \ours (with GPT-4o) after self-evolution.}
\label{fig:full_shortcuts}
\end{center}
\vskip -0.2in
\end{figure*}

\section{Full list of Self-Evolved Shortcuts}
\label{app:full_list_of_evolved_shortcuts}
Figure~\ref{fig:full_shortcuts} shows a full list of generated Shortcuts by \ours after self-evolution on all 25 tasks from \dataset benchmark.

\begin{figure*}[t]
\vskip 0.2in
\begin{center}
\centerline{\includegraphics[width=0.98\textwidth]{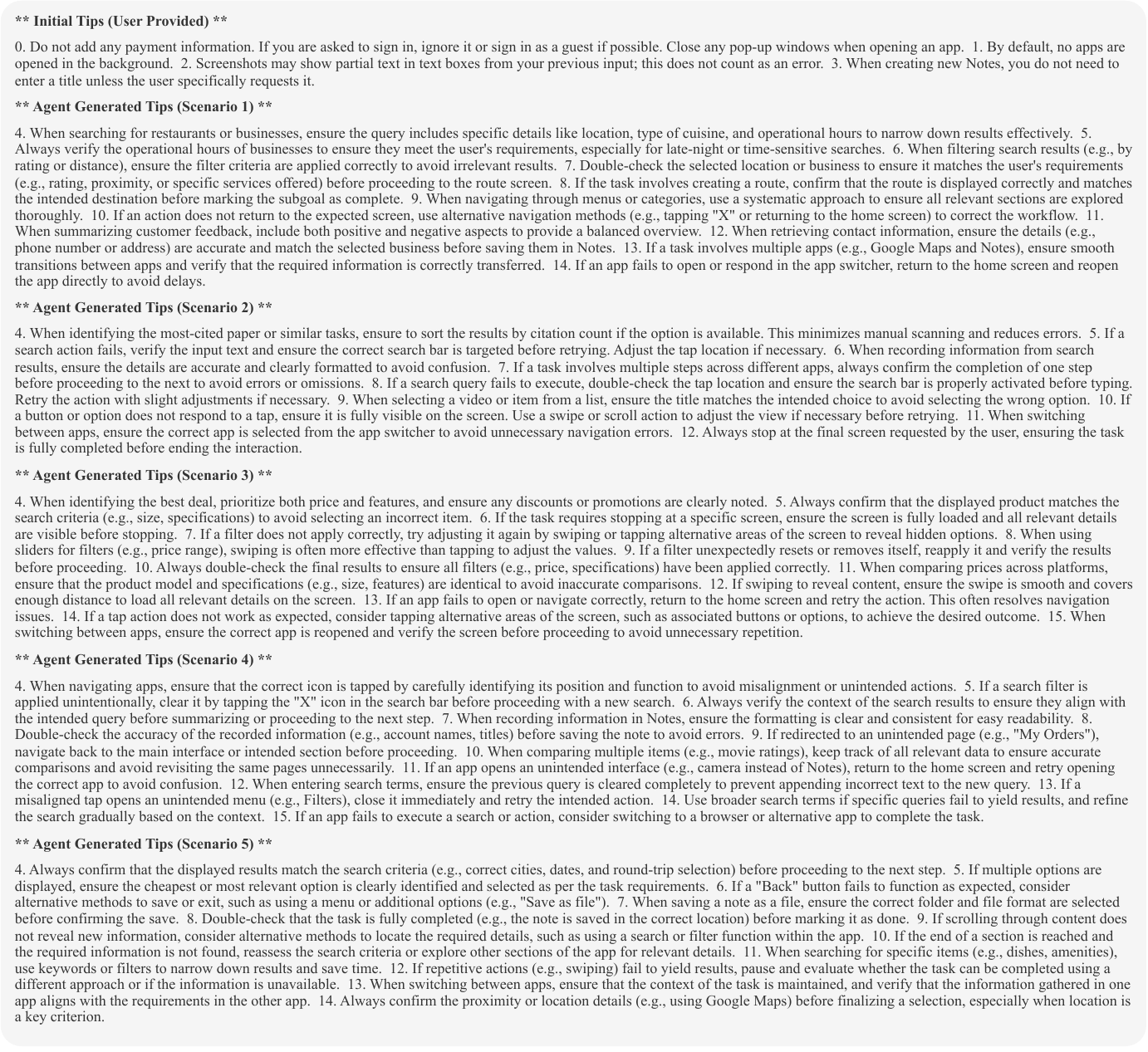}}
\caption{Full list of Tips generated by \ours (with GPT-4o) after self-evolution.}
\label{fig:full_tips}
\end{center}
\vskip -0.2in
\end{figure*}

\section{Full list of Self-Evolved Tips}
\label{app:full_list_of_evolved_tips}
Figure~\ref{fig:full_tips} shows a full list of generated Tips by \ours after self-evolution on all 25 tasks from \dataset benchmark.





\end{document}